\documentclass{adobe_research}
\usepackage[T1]{fontenc}
\usepackage{lmodern}

\usepackage{amsmath,amsfonts,bm}

\def\eqref#1{equation~\ref{#1}}
\def\1{\bm{1}}

\DeclareMathAlphabet{\mathsfit}{\encodingdefault}{\sfdefault}{m}{sl}
\SetMathAlphabet{\mathsfit}{bold}{\encodingdefault}{\sfdefault}{bx}{n}

\hypersetup{
    colorlinks=true,
    linkcolor=blue,
    citecolor=adobered,
    urlcolor=blue
}
\usepackage{url}
\usepackage{amsmath}
\usepackage{amssymb}
\usepackage{float}
\usepackage{colortbl}
\usepackage{enumitem}
\usepackage{wrapfig}
\usepackage{needspace}
\usepackage{capt-of}

\title{Adversarial Training for Pixel Diffusion}

\author[1,2*]{Xin Lin}
\author[2]{Zhifei Zhang}
\author[2]{Yuqian Zhou}
\author[2]{Haitian Zheng}
\authorbreak
\author[2]{Zhe Lin}
\author[3]{Ming-Hsuan Yang}
\author[1]{Truong Nguyen}

\affiliation[1]{UC San Diego}
\affiliation[2]{Adobe Research}
\affiliation[3]{UC Merced}
\contribution[*]{Work done during an internship at Adobe Research}

\newcommand{\up}{$\uparrow$}
\newcommand{\dn}{$\downarrow$}
\newcommand{\best}[1]{\textbf{#1}}

\begin{document}
\raggedbottom

\newcommand{\renderteaser}{%
\begin{figure}[H]
\centering
\setlength{\fboxsep}{0pt}
\newlength{\uw}\setlength{\uw}{\textwidth}
\newcommand{\tg}{\hspace{0.8pt}}
\newcommand{\lo}{\makebox[0.12\uw]{\scriptsize w/o GAN}}
\newcommand{\lw}{\makebox[0.12\uw]{\scriptsize w/ GAN}}
% --- Highlight: eight 512px DeCo pairs (our model), 2 rows x 4, flush to both edges ---
% column labels: rows 1 and 2 share the same 4-pair structure, so label only above row 1
\noindent\makebox[\uw]{\lo\tg\lw\hfill\lo\tg\lw\hfill\lo\tg\lw\hfill\lo\tg\lw}\\[0.6pt]
\noindent\makebox[\uw]{%
\includegraphics[width=0.12\uw]{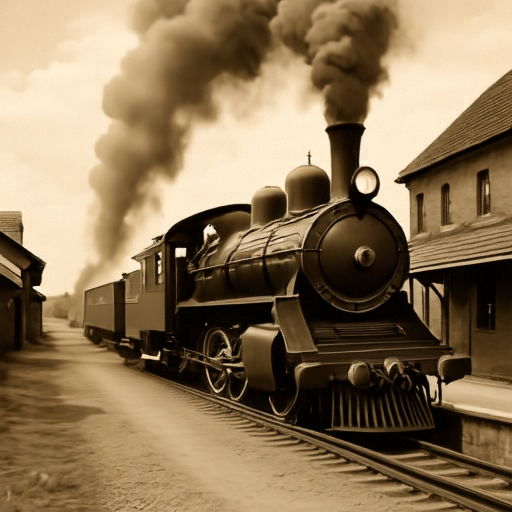}\tg\includegraphics[width=0.12\uw]{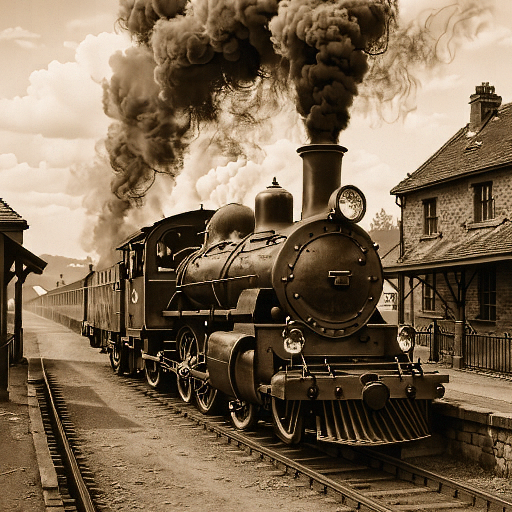}\hfill%
\includegraphics[width=0.12\uw]{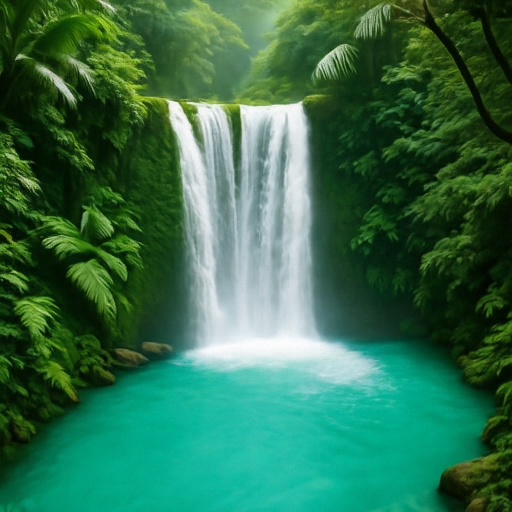}\tg\includegraphics[width=0.12\uw]{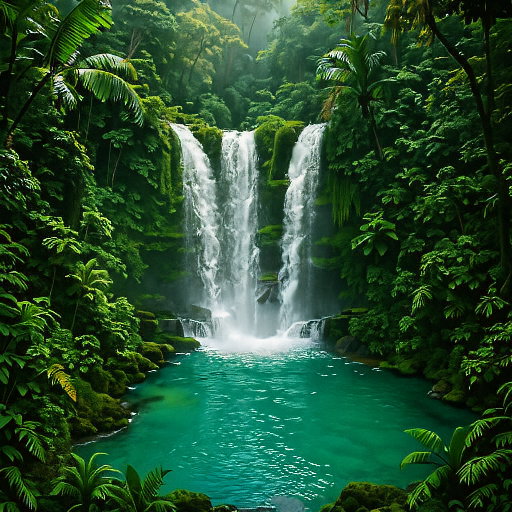}\hfill%
\includegraphics[width=0.12\uw]{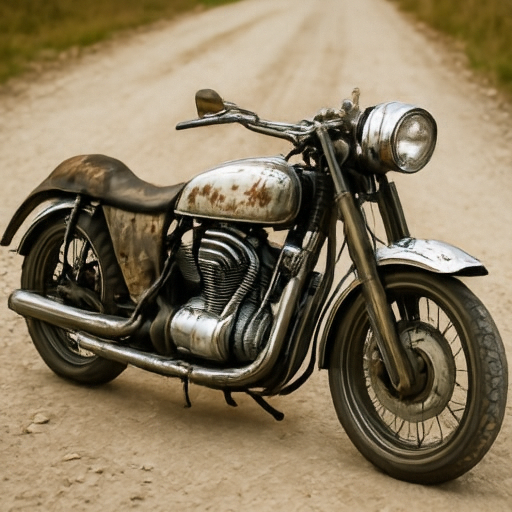}\tg\includegraphics[width=0.12\uw]{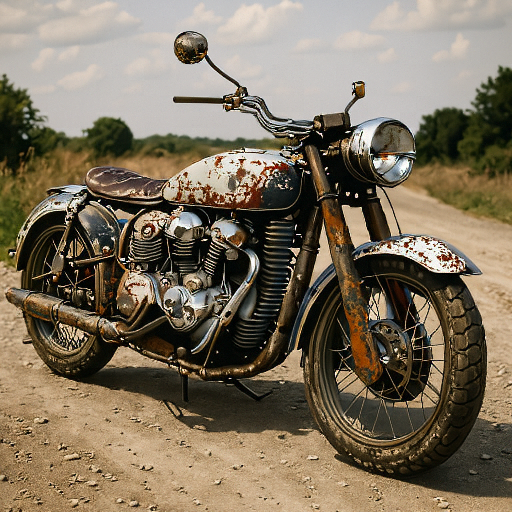}\hfill%
\includegraphics[width=0.12\uw]{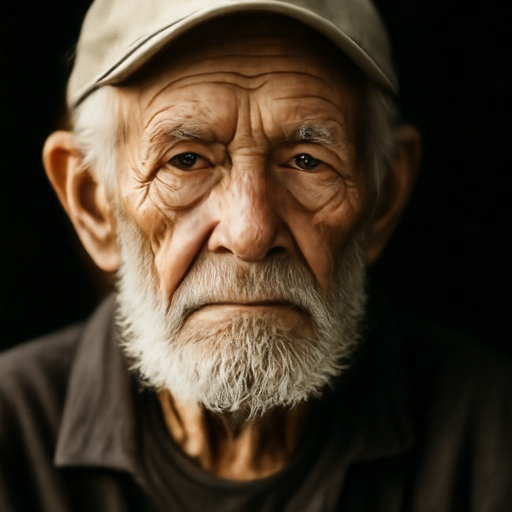}\tg\includegraphics[width=0.12\uw]{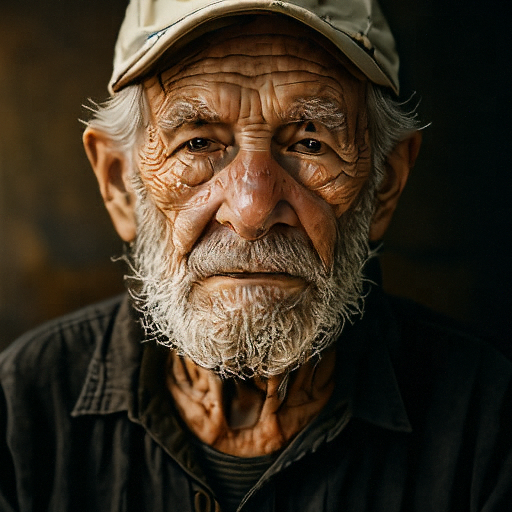}%
}\\[1.2pt]
\noindent\makebox[\uw]{%
\includegraphics[width=0.12\uw]{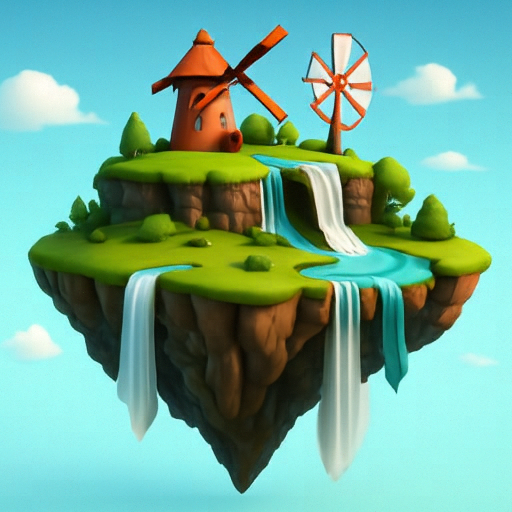}\tg\includegraphics[width=0.12\uw]{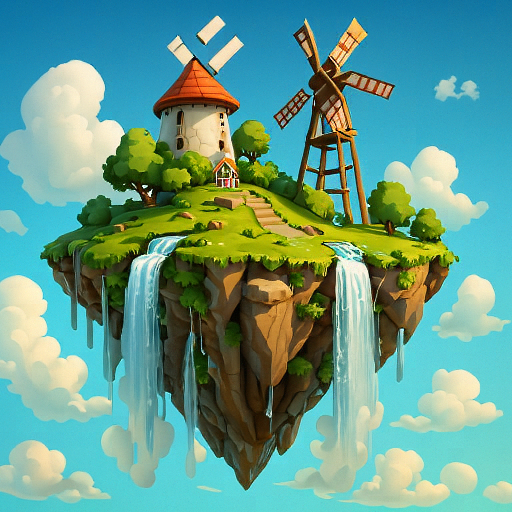}\hfill%
\includegraphics[width=0.12\uw]{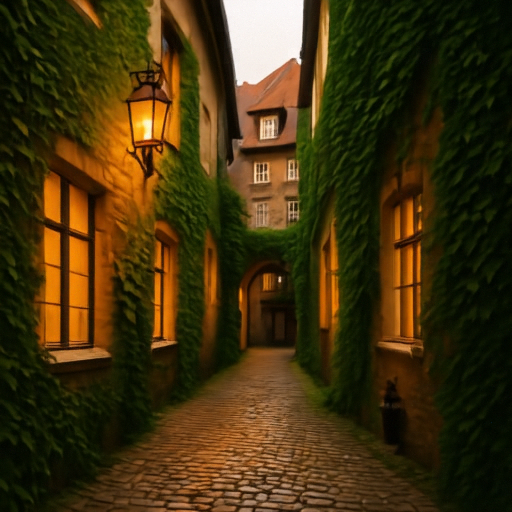}\tg\includegraphics[width=0.12\uw]{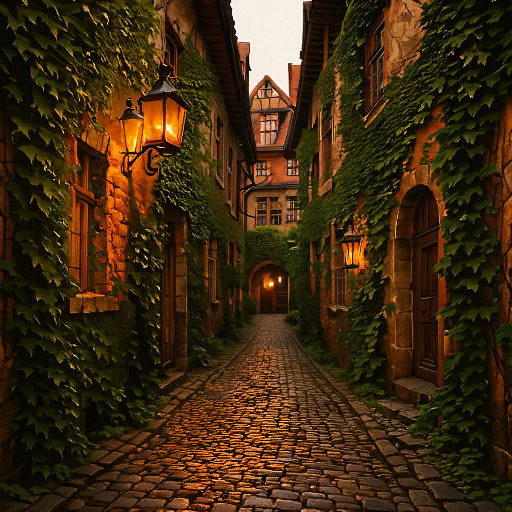}\hfill%
\includegraphics[width=0.12\uw]{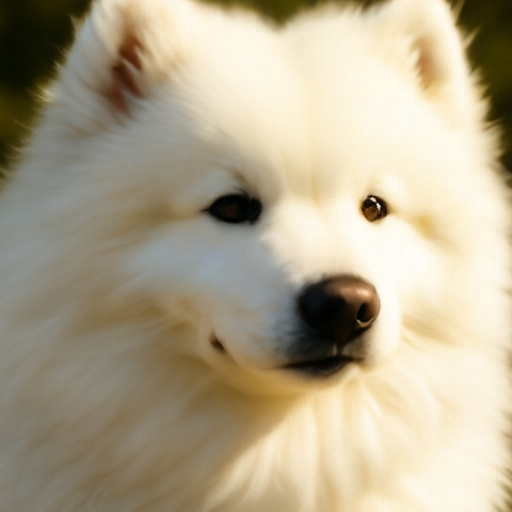}\tg\includegraphics[width=0.12\uw]{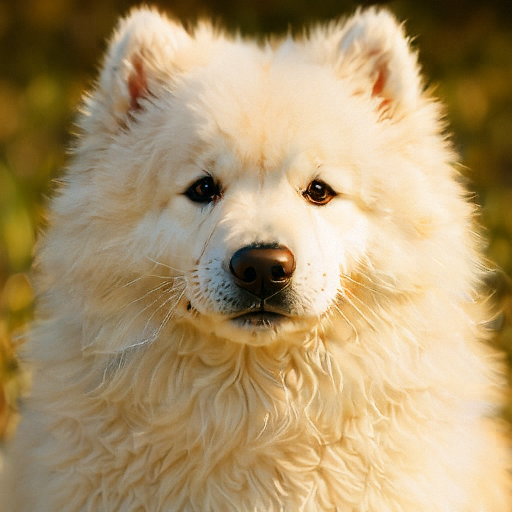}\hfill%
\includegraphics[width=0.12\uw]{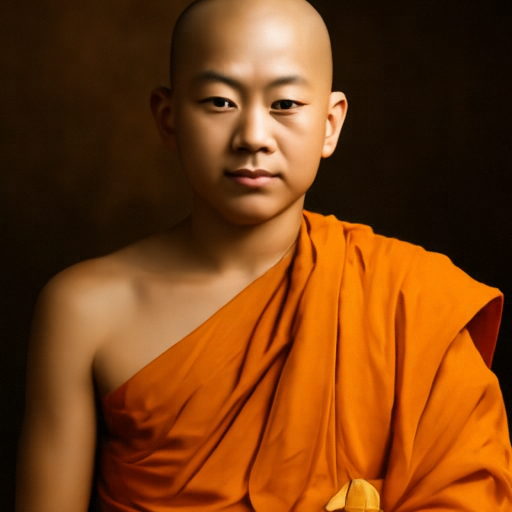}\tg\includegraphics[width=0.12\uw]{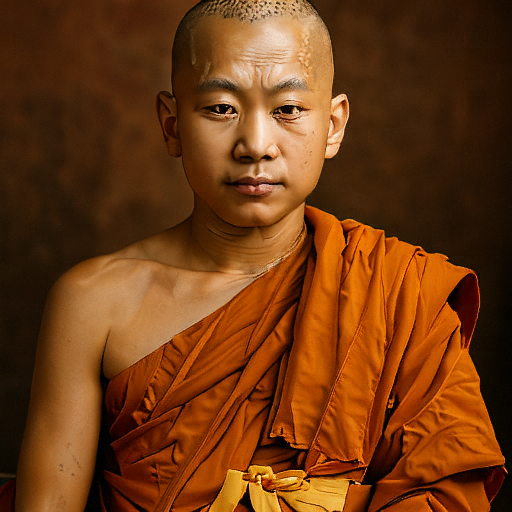}%
}\\[1.6pt]
% --- Two 1K pairs (white tiger + cartoon) with zoom-in insets, flush to both edges ---
\noindent\makebox[\uw]{%
\includegraphics[width=0.144\uw]{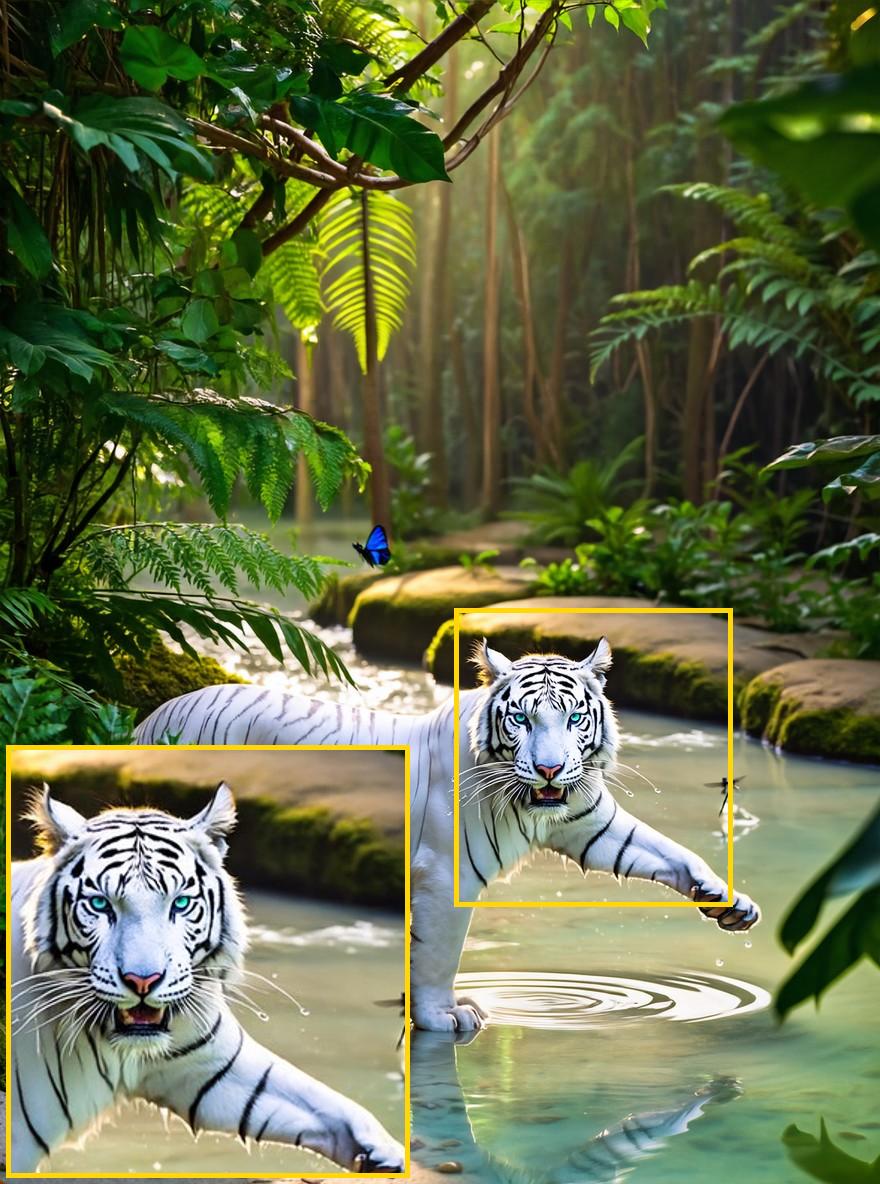}\tg\includegraphics[width=0.144\uw]{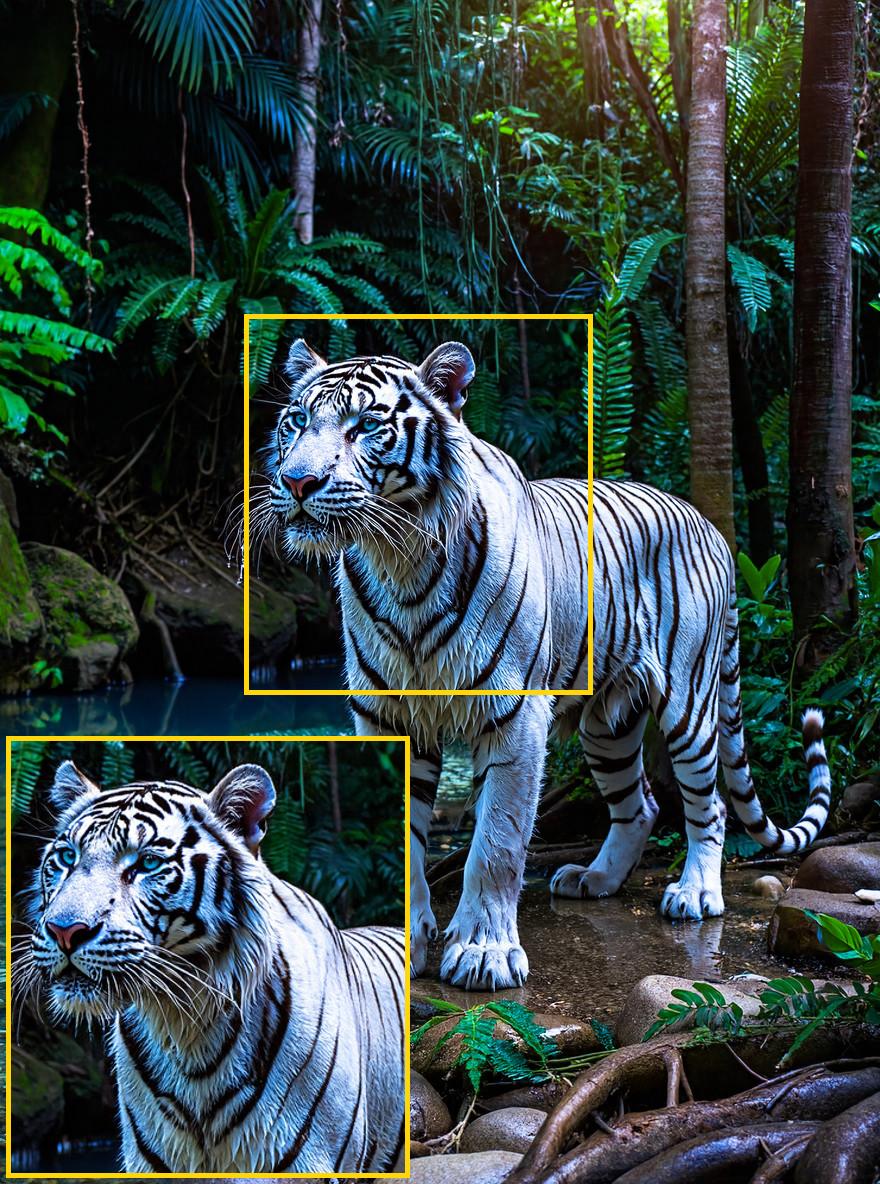}\hfill%
\includegraphics[width=0.343\uw]{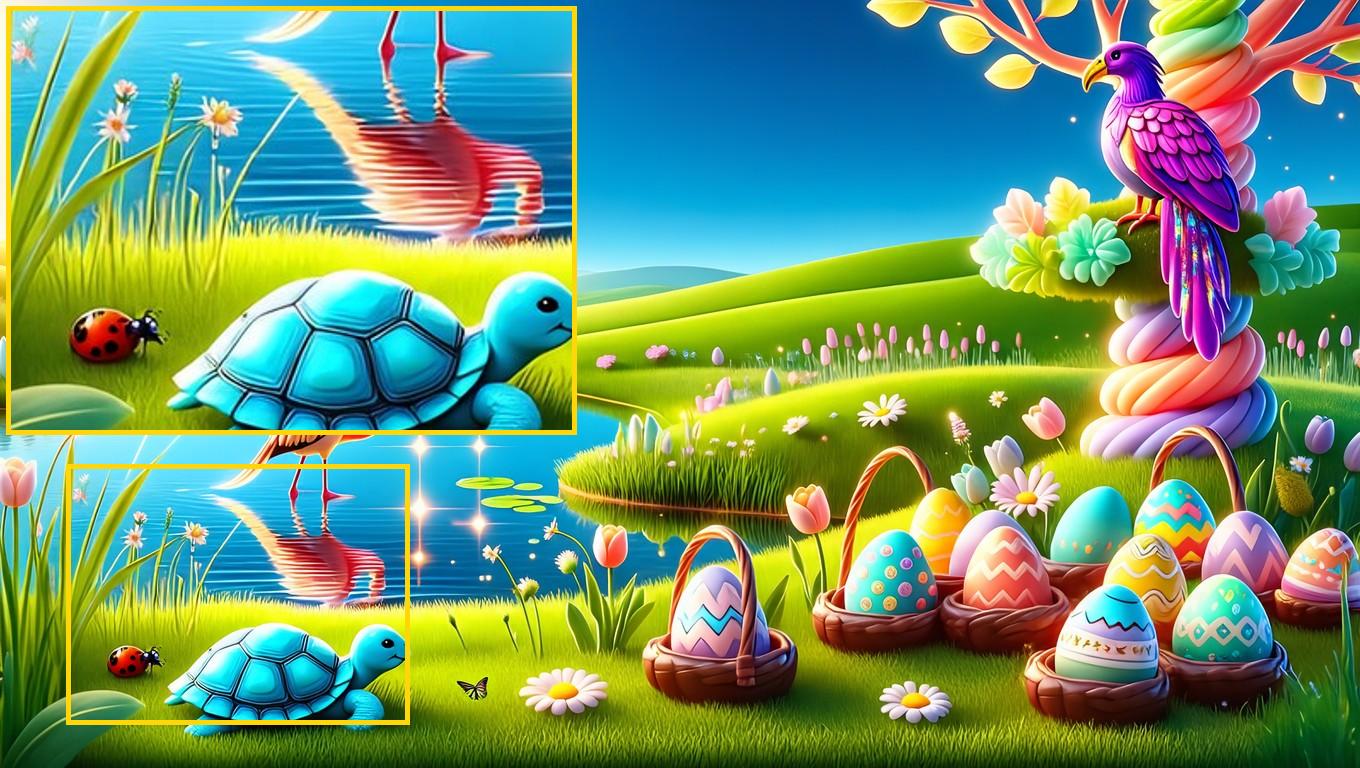}\tg\includegraphics[width=0.343\uw]{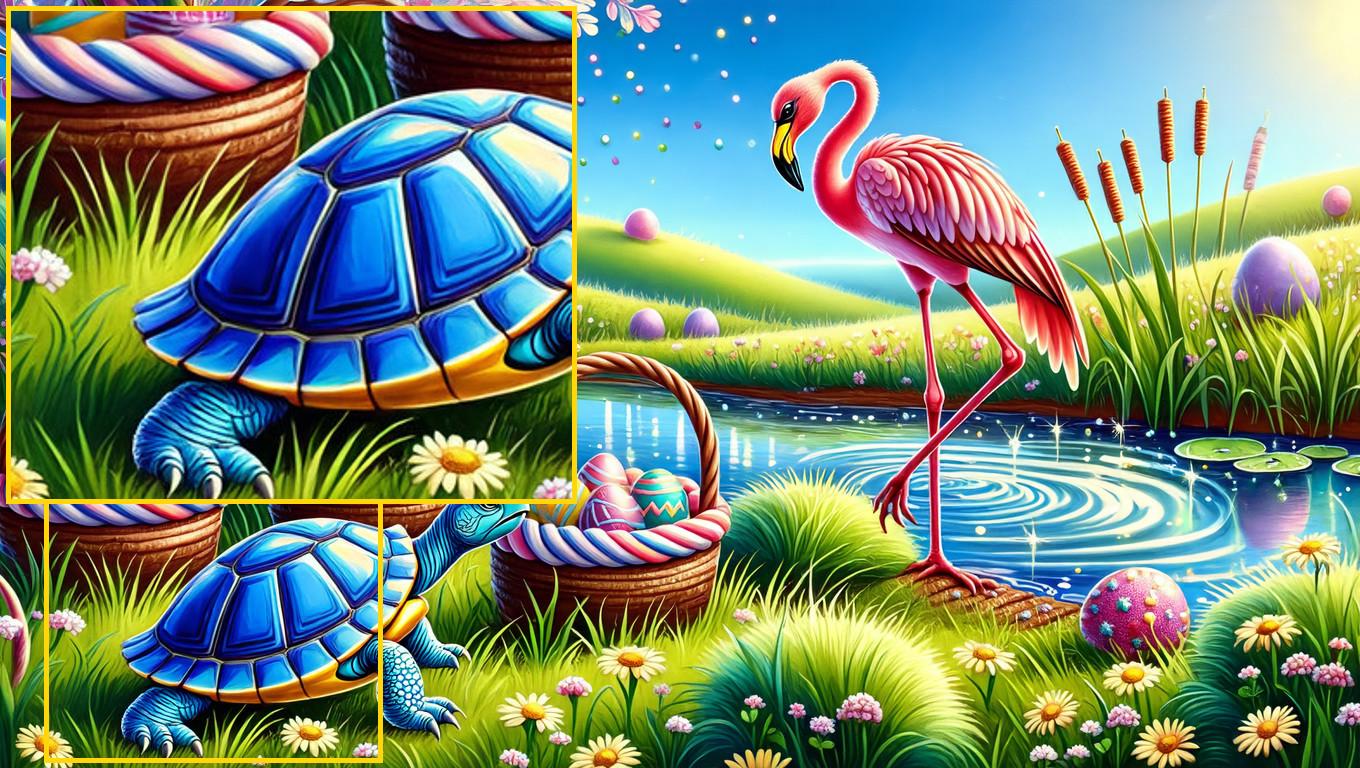}%
}\\[0.6pt]
\noindent\makebox[\uw]{\makebox[0.144\uw]{\scriptsize w/o GAN}\tg\makebox[0.144\uw]{\scriptsize w/ GAN}\hfill\makebox[0.343\uw]{\scriptsize w/o GAN}\tg\makebox[0.343\uw]{\scriptsize w/ GAN}}\\[2pt]
\caption{Each pair shows w/o GAN (left) and w/ GAN (right), where w/ GAN means adding the adversarial loss during post-training: eight $512^2$ pairs across diverse prompts and styles (top) and two 1K pairs with zoom-ins (bottom).}
\label{fig:teaser}
\end{figure}
}

\begin{adobeabstract}
Pixel diffusion models generate RGB images directly, avoiding the bottleneck of an autoencoder, yet their outputs still systematically underrepresent fine-scale natural-image statistics.
We show that adversarial learning provides an effective post-training correction for this deficiency. Starting from a pretrained model, we retain its original diffusion or flow-matching objective and add an adversarial loss to the predicted output at non-high-noise timesteps, leaving the model architecture and sampling procedure unchanged.
To our knowledge, this is the first systematic study of adversarial post-training for pixel diffusion. 
Across two pixel backbones, the method jointly improves distribution fidelity, coverage, prompt alignment, and perceptual quality. 
We further investigate why it works. Frequency-band and power-law analyses show that the original models systematically underproduce natural-image high-frequency content, while adversarial post-training restores this missing spectral power. 
In contrast, perceptual loss also increases high-frequency content but sacrifices distribution fidelity and prompt alignment. Nearest-neighbor, recall, and matched no-GAN SFT controls further rule out memorization, mode dropping, and additional optimization as simple explanations.
Finally, we examine the boundary of this effect. Under the tested latent diffusion configurations, the same procedure does not produce comparable
joint gains and adds almost no decoded high-frequency power.  
These results identify direct output access to the image statistics being corrected as a key factor governing when adversarial post-training succeeds. 
\end{adobeabstract}

\renderteaser

\section{Introduction}

Recent work has renewed interest in pixel diffusion for high-resolution text-to-image (T2I)
generation \citep{hoogeboom2023simplediffusion,chen2023pixeldiffusion,ma2026deco,ma2026pixelgen}. Unlike
latent diffusion, which generates a compressed representation and relies on a separately trained decoder to
render RGB \citep{rombach2022ldm,podell2023sdxl,chen2024pixart,xie2024sana}, a pixel model predicts the final
image. This direct parameterization removes the autoencoding bottleneck, but also makes the denoiser
responsible for generating both global structure and fine image detail. Converged pixel models capture text
semantics and coarse composition well, yet still underrepresent fine-scale image statistics \citep{ma2026pixelgen, ma2026deco}. 
This gap is
visible in the smooth, under-textured outputs in Figure~\ref{fig:teaser} and measurable in the
radial power spectrum: the DeCo \citep{ma2026deco} baseline exhibits a substantially steeper slope than real COCO images
($\alpha=2.59$ vs.\ $\approx2.19$; Table~\ref{tab:spectrum-app}), indicating a measurable high-frequency
deficit. We ask whether adversarial post-training can correct this residual detail gap without trading away
distribution fidelity, diversity, or prompt alignment.

Adversarial objectives align a generator's distribution with real data
\citep{goodfellow2014gan, lin2023unsupervised, lin2025re}. Within diffusion systems, they are also used to pretrain the separately trained
VAE/autoencoder decoder that maps latent codes to RGB, mitigating the blur of reconstruction-only training
\citep{esser2021vqgan,rombach2022ldm}. When applied to diffusion generators, adversarial or
distribution-matching objectives have mainly been used to enable large denoising transitions or distill
models to one or a few sampling steps
\citep{xiao2022ddgan,sauer2024add,yin2024dmd,yin2024dmd2}. We study a different role:
\emph{adversarial post-training} as a pure quality correction for an already-converged, multi-step pixel
diffusion model. We retain its original objective and add a hinge adversarial loss on the predicted clean
image $\hat{x}_0$, excluding high-noise timesteps where global structure is not yet reliable.
Throughout, we denote this adversarial-loss addition as \emph{+GAN} (or \emph{w/ GAN} in figures).
The procedure uses no preference labels or reward model, performs no distillation, and does not reduce the
number of sampling steps. Across DeCo \cite{ma2026deco} and PixelGen \cite{ma2026pixelgen}, it jointly improves distribution fidelity, coverage,
prompt alignment, and no-reference image quality (Table~\ref{tab:c1main}). On DeCo, the DINOv2-text configuration improves FID from $33.27$ to $28.59$, recall from $0.361$ to $0.406$, and DPG Score~\citep{hu2024dpgbench} from $81.6$ to
$83.3$.

We trace this gain to a systematic residual error in the pixel models' outputs. Both baselines underproduce
natural-image high-frequency (HF) statistics; adversarial post-training restores that missing power and, on
DeCo, moves the radial power-spectrum slope from $\alpha=2.59$ to $2.24$, close to real images at
approximately $2.19$. The effect is not generic sharpening. Non-adversarial perceptual supervision offers another route to sharper pixel diffusion, as explored by PixelGen with LPIPS/DINO features
\citep{ma2026pixelgen}. In our matched comparison, this perceptual objective also adds HF, but induces a desaturated, low-contrast domain shift and degrades FID and prompt alignment, whereas
the GAN improves distributional and perceptual quality together. Recall increases, DINOv2 nearest-neighbor
similarity to the training set is unchanged (Table~\ref{tab:nn-memorization}), and all gains are measured
against matched-step no-GAN SFT controls (Table~\ref{tab:c1main}), ruling out mode dropping, memorization,
and additional optimization as simple explanations.

To explain when adversarial refinement can realize this gain, we adopt a two-condition view: it requires
(i) a correctable error in an output subspace and (ii) sufficient local access from the trainable output to
that subspace. Pixel diffusion satisfies both conditions: its output is HF-deficient RGB, and no decoder
intervenes between the trainable prediction, the discriminator, and the final pixels. Two tested latent
models provide a mechanism-matched output-access comparison. For each model, the GAN run is paired with its
own no-GAN control under aligned training and evaluation. PixArt-$\alpha$ and SANA show no comparable joint
improvement, while a direct perturbation probe through the frozen PixArt VAE measures a
$3.5\text{--}11\times$ weaker decoded-HF response. This matched comparison and direct probe identify limited
decoded-HF access as an important mechanism; discriminator, weight, and noise-gate ablations separately show
how design choices shift the empirical metric trade-offs.

We organize our contributions around three questions---whether adversarial post-training works, why it
works, and when it works:
\begin{itemize}
\item We propose \emph{adversarial post-training} as a quality-refinement approach for pretrained T2I pixel
diffusion models. To our knowledge, we are the first to systematically study GAN-based post-training for
pixel diffusion. Across two pixel backbones, it jointly improves distribution fidelity, coverage, prompt
alignment, and perceptual quality without changing the model architecture or inference procedure.
\item We comprehensively analyze why adversarial post-training works in pixel space. Frequency-band and
power-law analyses reveal a systematic deficit in natural-image HF statistics and show that the GAN restores
the missing power. A matched perceptual-loss comparison further distinguishes this correction from
generic sharpening: both objectives add HF, but only the GAN improves distributional and perceptual quality
together.
\item We characterize when adversarial refinement works through matched pixel--latent comparisons: direct pixel diffusion models show consistent joint gains, whereas latent diffusion models do not.
Discriminator, noise-gate, and adversarial-weight ablations further map the operating
conditions and metric trade-offs within pixel diffusion.
\end{itemize}

\section{Related work}

\paragraph{Generative adversarial networks.} GANs \citep{goodfellow2014gan} long defined the state of the
art in image synthesis, from the StyleGAN family \citep{karras2019stylegan,karras2020stylegan2} to
conditional image-to-image models built on patch discriminators \citep{isola2017pix2pix}. Scaled-up
text-to-image GANs \citep{sauer2023stylegant,kang2023gigagan} remain competitive on sharpness and sampling
speed, but trail diffusion on sample diversity and prompt controllability. A recurring lesson is that the
\emph{discriminator} governs what a GAN can learn. Projected and feature-space discriminators built on
\emph{frozen} pretrained backbones \citep{sauer2021projectedgan,sauer2023stylegant} markedly stabilize and
strengthen training, while augmentation schemes such as DiffAugment and adaptive discriminator
augmentation curb discriminator overfitting on limited data \citep{zhao2020diffaugment,karras2020ada}. We build directly on this line, using frozen DINOv2/DINOv3/SigLIP backbones \citep{oquab2024dinov2,simeoni2025dinov3,zhai2023siglip} and text-conditioned projection heads \citep{sauer2023stylegant} as our discriminators throughout.

\paragraph{Adversarial losses in diffusion.} Although diffusion models are trained by denoising
\citep{ho2020ddpm,song2021scoresde}, an adversarial term still appears at two main points in the modern T2I
stack. First, in \emph{tokenizer training}: the VAE/autoencoder underlying latent diffusion is trained with
a combined perceptual (LPIPS) and patch-GAN objective \citep{esser2021vqgan,rombach2022ldm}, so the
discriminator acts on the \emph{decoder} that renders pixels rather than on the diffusion model itself.
Second, in \emph{distillation}: a discriminator lets a one- or few-step student match the data
distribution, as in adversarial diffusion distillation and distribution-matching distillation
\citep{sauer2024add,sauer2024ladd,yin2024dmd,yin2024dmd2}. In both cases the GAN serves tokenization or step
reduction. In contrast, we isolate the GAN as a pure multi-step quality term for an already-converged
pixel-diffusion model, changing neither its architecture nor its number of sampling steps.

\paragraph{Pixel diffusion.} Pixel diffusion models denoise directly on RGB pixels
\citep{ho2020ddpm,dhariwal2021adm,nichol2021iddpm}, so the network outputs the final image and must
synthesize all of its detail itself. Modern backbones adopt transformer denoisers \citep{peebles2023dit}
and flow-matching or improved diffusion formulations \citep{lipman2023flowmatching,karras2024edm2}. Once
restricted to low resolution or multi-stage cascades, pixel diffusion has recently re-emerged as a
competitive single-stage paradigm: efficient high-resolution designs
\citep{hoogeboom2023simplediffusion,chen2023pixeldiffusion}, flow- and neural-field variants
\citep{chen2025pixelflow,wang2025pixnerd}, and transformer backbones that decouple global structure from
local detail \citep{chen2025dip,yu2025pixeldit}, alongside strong text-to-image models such as DeCo
\citep{ma2026deco} and PixelGen \citep{ma2026pixelgen}, which we adopt as our backbones.

\section{Setup}
\label{sec:setup}

\begin{figure}[t]
\centering
\newlength{\ccw}\setlength{\ccw}{0.233\linewidth}
\definecolor{ngbar}{HTML}{9AA0A6}\definecolor{ganbar}{HTML}{1D7268}
\newcommand{\chdr}[2]{\colorbox{#1}{\parbox[c][1.5ex][c]{\ccw}{\centering\scriptsize\textcolor{white}{\textbf{#2}}}}}
\newcommand{\cim}[1]{\includegraphics[width=\ccw]{figs/#1}}
\setlength{\tabcolsep}{1.2pt}
\renewcommand{\arraystretch}{0.4}\setlength{\fboxsep}{0pt}
\begin{tabular}{cccc}
\chdr{ngbar}{w/o GAN} & \chdr{ganbar}{w GAN} & \chdr{ngbar}{w/o GAN} & \chdr{ganbar}{w GAN}\\
\cim{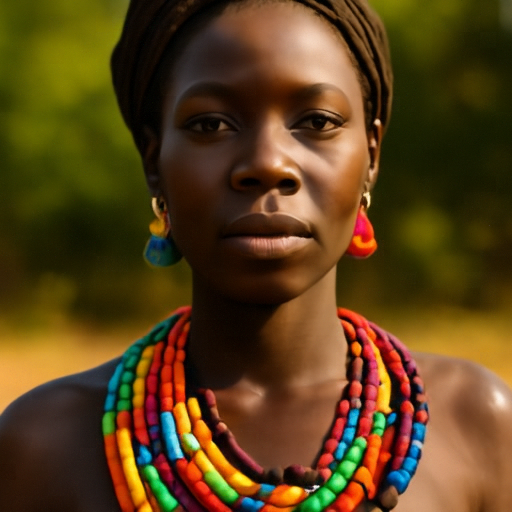} & \cim{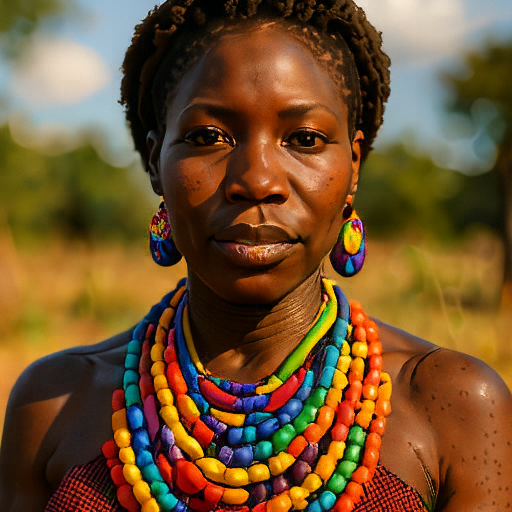} & \cim{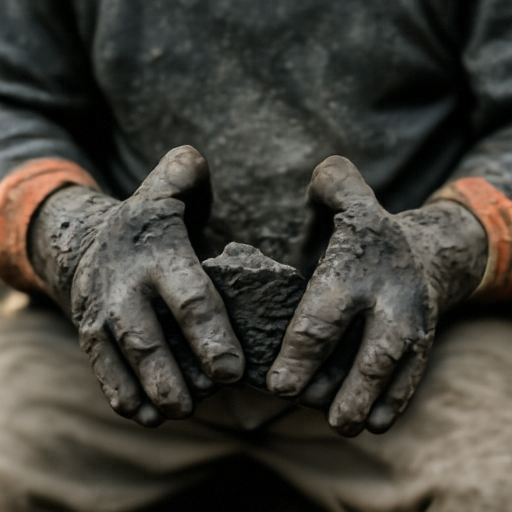} & \cim{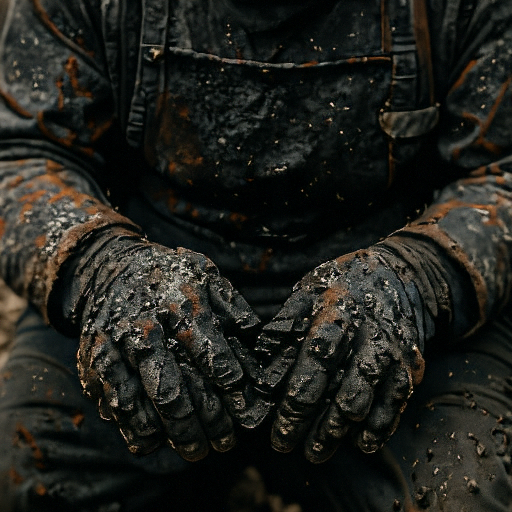}\\[2.5pt]
\chdr{ngbar}{w/o GAN} & \chdr{ganbar}{w GAN} & \chdr{ngbar}{w/o GAN} & \chdr{ganbar}{w GAN}\\
\cim{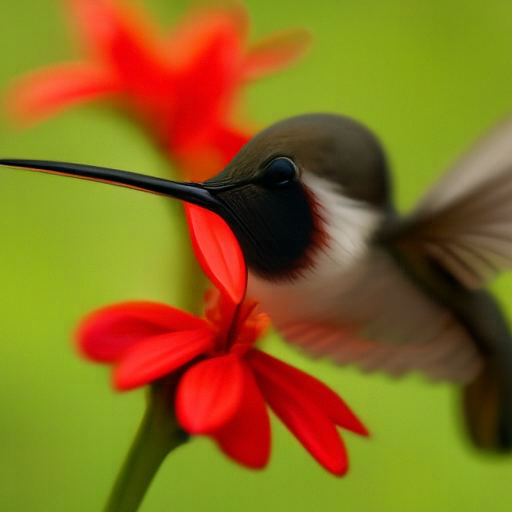} & \cim{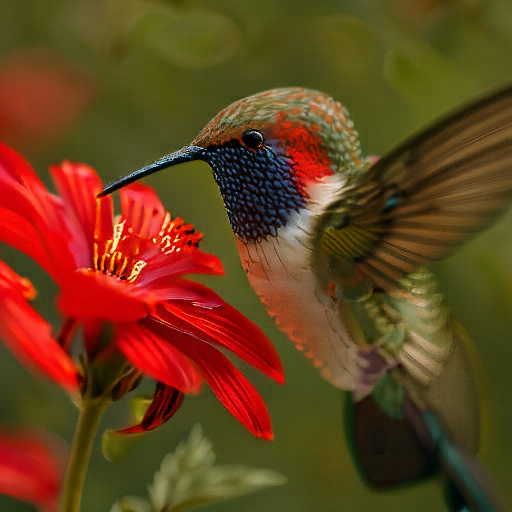} & \cim{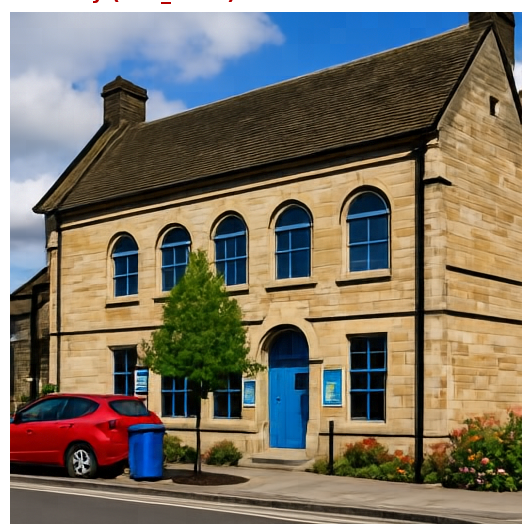} & \cim{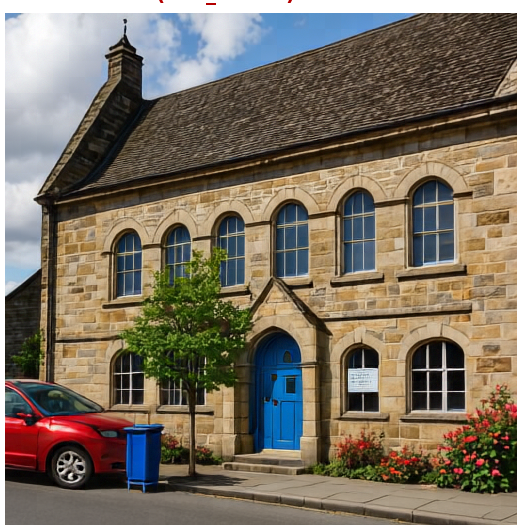}\\
\end{tabular}
\caption{Same-prompt, same-seed comparison with and without GAN fine-tuning.}
\label{fig:c1-qual}
\vspace{-5pt}
\end{figure}

\paragraph{Adversarial post-training.} We treat the adversarial loss as \emph{post-training}: from a
converged model we add an adversarial/GAN loss to the original diffusion/flow-matching objective and
continue training. The generator sees the standard diffusion/flow-matching
loss plus $\mathcal{L}_G=w_\mathrm{gan}\mathbb{E}[-D(h(\hat{y}_0))]$, while the discriminator minimizes
$\mathcal{L}_D=\tfrac12\mathbb{E}[\mathrm{relu}(1-D(h(y_0)))]+
\tfrac12\mathbb{E}[\mathrm{relu}(1+D(h(\hat{y}_0)))]$.
Here $y_0$ denotes the clean target in the model's native output space---RGB $x_0$ for DeCo and PixelGen,
and latent $z_0$ for SANA and PixArt-$\alpha$---and $h$ maps that space to the discriminator input.
For velocity-prediction DeCo and SANA, $y_t=(1-\sigma_t)y_0+\sigma_t\epsilon$ and
$\hat{y}_0=y_t-\sigma_t v_\theta(y_t,t)$. PixelGen directly predicts
$\hat{x}_0=x_\theta(x_t,t)$; eps-prediction PixArt-$\alpha$ uses
$\hat{z}_0=(z_t-\sqrt{1-\bar\alpha_t}\,\epsilon_\theta(z_t,t))/\sqrt{\bar\alpha_t}$.
For the pixel models $h$ is the identity; decoded-RGB latent variants use the frozen VAE decoder,
whereas the other latent ablations use native features as described in \S\ref{sec:output-comparison}.

\paragraph{Backbone and data.} We study two pixel models, DeCo~\citep{ma2026deco} and
PixelGen~\citep{ma2026pixelgen}, and two latent models, PixArt-$\alpha$~\citep{chen2024pixart} and
SANA~\citep{xie2024sana}. All quantitative models are fine-tuned from public checkpoints on
BLIP3o-60k \cite{chen2025blip3o}, the same dataset originally used to train DeCo and PixelGen. For each backbone, the GAN and no-GAN controls share the same starting checkpoint, data, and training horizon. Please refer to Appendix~\ref{sec:supp} for details of the different discriminator architectures, model and training details.

\paragraph{Timestep gating.} We apply the adversarial loss at non-high-noise (sufficient-SNR) timesteps using the signal-fraction gate $\alpha_t=1-\sigma_t\ge\tau$, since $\hat{x}_0$ is not yet a meaningful image in the high-noise regime. Exact gate choices, training details, and the discriminators are described in
Appendix~\ref{sec:supp}. The gate thus excludes samples whose global layout is still unresolved while retaining the stage at which local appearance can be refined.

\enlargethispage{\baselineskip}
\paragraph{Evaluation.} We report metrics in four groups.
\emph{(i) Prompt alignment:} DPG Score on DPG-Bench \citep{hu2024dpgbench}, which measures how faithfully the image follows the text. \emph{(ii) Distribution fidelity and diversity} on COCO-30k \citep{lin2014coco}: FID
\citep{heusel2017fid} and patch-FID (pFID) for feature-distribution distance at the image and patch scale,
CMMD \citep{jayasumana2024cmmd} as a lower-bias alternative to FID, IS \citep{salimans2016is} for
quality/diversity, recall \citep{kynkaanniemi2019precisionrecall} for mode coverage, and CLIP score
\citep{radford2021clip} for image--text agreement. \emph{(iii) No-reference image quality}, scoring a
single image with no ground-truth reference: TOPIQ, MUSIQ, MANIQA, and NIQE
\citep{chen2024topiq,ke2021musiq,yang2022maniqa,mittal2013niqe}. \emph{(iv) Naturalness of spatial
statistics:} the radial power spectrum and its fitted power-law slope $\alpha$ (natural images have
$\alpha\!\approx\!2$ \citep{ruderman1994natural,vanderschaaf1996powerspectra}), which reveals whether
high-frequency content matches natural images rather than being over- or under-sharpened. 

\section{Adversarial post-training improves pixel diffusion}
\label{sec:c1}

\subsection{Joint gains across pixel backbones}

Adversarial post-training reliably improves pixel diffusion for T2I on both backbones (Table~\ref{tab:c1main}). On DeCo, it improves the evaluation battery across DPG-Bench and COCO-30k: FID
$33.3\!\to\!28.6$, pFID $27.9\!\to\!24.4$, CMMD $0.836\!\to\!0.736$, recall $0.36\!\to\!0.41$, TOPIQ
$0.71\!\to\!0.77$, MANIQA $0.64\!\to\!0.71$, and DPG Score $81.6\!\to\!83.3$. The same GAN also improves
PixelGen on every axis. Figure~\ref{fig:c1-qual}
shows the same effect qualitatively: the GAN adds fine detail while preserving global structure and color.
% We next characterize this output change and rule out generic sharpening and other trivial explanations; \S\ref{sec:c2} then tests when and why the refinement works.

\begin{table}[t]
\caption{\textbf{GAN vs.\ no-GAN on two pixel backbones.} DPG Score is evaluated on DPG-Bench~\citep{hu2024dpgbench}; all other metrics use COCO-30k~\citep{lin2014coco}. \best{Bold} $=$ better within each model.}
\label{tab:c1main}
\begin{center}\footnotesize
\setlength{\tabcolsep}{3.5pt}
\resizebox{\textwidth}{!}{%
\begin{tabular}{llccccccccc}
\toprule
 & variant & DPG Score\,\up & FID\,\dn & IS\,\up & CMMD\,\dn & Rec\,\up & pFID\,\dn & TOPIQ\,\up & MUSIQ\,\up & MANIQA\,\up \\
\midrule
\multirow{2}{*}{\textbf{DeCo} (pixel)}
  & no-GAN SFT & 81.6 & 33.27 & 39.35 & 0.836 & 0.361 & 27.91 & 0.711 & 75.5 & 0.636 \\
  & \;+GAN & \best{83.3} & \best{28.59} & \best{39.77} & \best{0.736} & \best{0.406} & \best{24.38} & \best{0.768} & \best{76.5} & \best{0.712} \\
\midrule
\multirow{2}{*}{\textbf{PixelGen} (pixel)}
  & no-GAN SFT & 78.4 & 33.94 & 38.58 & 0.762 & 0.319 & 30.81 & 0.755 & 75.7 & 0.645 \\
  & \;+GAN & \best{80.8} & \best{33.20} & \best{38.76} & \best{0.725} & \best{0.403} & \best{30.51} & \best{0.799} & \best{77.0} & \best{0.723} \\
\bottomrule
\end{tabular}}
\end{center}
\end{table}

\subsection{The GAN restores missing natural-image high frequencies}

\begin{wrapfigure}{r}{0.44\linewidth}
\vspace{-15pt}
\centering
\includegraphics[width=0.96\linewidth,trim=0 0 278 0,clip]{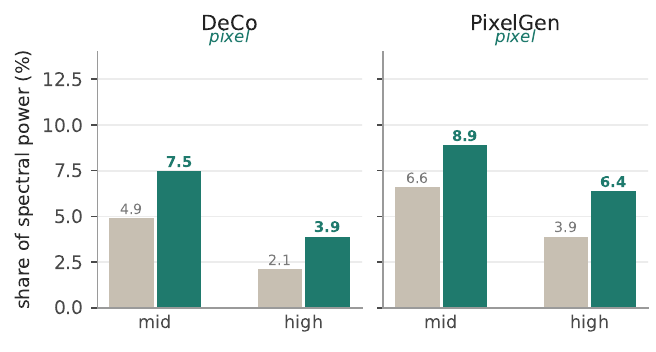}
\vspace{-7pt}
\caption{Pixel radial-profile band share (\%) over $30{,}000$ images/model; gray $=$ no-GAN, green $=+$GAN.}
\label{fig:hfbands}
\vspace{-8pt}
\end{wrapfigure}

We inspect the frequency composition: the radial-profile band share
$|\hat{I}(f)|^2$ in the mid and high bands. Mid frequency is
$0.10\!\le\!f\!\le\!0.25$ cyc/px and high frequency is $0.25\!<\!f\!\le\!0.50$ cyc/px, with
$0.50$ cyc/px the Nyquist limit. Each bar in Figure~\ref{fig:hfbands} is the radial-profile
power in that interval divided by the total radial-profile power, averaged over $30{,}000$ images per model. On both pixel
backbones, the GAN moves a substantial share of power into these bands: DeCo mid $4.9\%\!\to\!7.5\%$, high
$2.1\%\!\to\!3.9\%$; PixelGen mid $6.6\%\!\to\!8.9\%$, high $3.9\%\!\to\!6.4\%$. To determine whether the
added spectral power is \emph{natural} detail rather than noise, we measure the
\emph{radial power spectrum} of the generations following the
azimuthally-averaged power-spectrum method of \citet{koch2010fourier}, applied to deep-network generations
as in \citet{dzanic2020fourier}. For each image we take the luminance channel, apply a 2-D Hann window,
take the 2-D FFT, and azimuthally average the squared magnitude $|\hat{I}(f)|^2$ over rings of constant
spatial frequency $f$; averaging over $30{,}000$ images yields one power-vs-frequency curve per model, whose
log--log slope we fit by least squares.

\begin{wraptable}{r}{0.42\linewidth}
\vspace{-15pt}
\centering\small
\caption{Pixel spectral statistics on COCO-30k; natural $\alpha\!\approx\!2.19$.}
\label{tab:spectrum-app}
\setlength{\tabcolsep}{4pt}
\resizebox{0.98\linewidth}{!}{%
\begin{tabular}{lccc}
\toprule
model & HF log-$\Delta$ & $\alpha$ w/o GAN & $\alpha$ $+$GAN \\
\midrule
DeCo & \best{$+0.34$} & 2.59 & \best{2.24} \\
\emph{real} & -- & \multicolumn{2}{c}{$\approx 2.19$} \\
\bottomrule
\end{tabular}}
\vspace{-8pt}
\end{wraptable}

Natural images obey a power law $P(f)\propto f^{-\alpha}$ with $\alpha\approx2$
\citep{ruderman1994natural,vanderschaaf1996powerspectra}: a larger $\alpha$ means power decays too fast
with frequency, so the image is deficient in high frequency (soft, blurry), while $\alpha$ near the
natural value means fine-scale detail matches real images. We summarize each model by two numbers: the
fitted slope $\alpha$, and the change $\Delta$ in high-frequency band power ($f>0.25$ cyc/px) between the
GAN and no-GAN model, in dex ($\log_{10}$). Unlike the normalized band-power shares in
Figure~\ref{fig:hfbands}, HF log-$\Delta$ measures the change in unnormalized high-band log-power
($+$GAN minus w/o GAN). On DeCo, the GAN raises the HF band by
$+0.34$ dex and pulls the slope from an HF-deficient $\alpha\!=\!2.59$ (no-GAN) toward the natural
law $\alpha\!=\!2.24$ (real COCO $\approx2.19$), \emph{without} flattening toward white noise (which would
drive $\alpha\!\to\!0$; Table~\ref{tab:spectrum-app}). This is where the pixel model's ``added detail''
is: it puts energy back into the frequencies a converged diffusion model under-produces. The
added energy is coherent texture, not artifacts.

\subsection{The gain is not generic sharpening or mode dropping}

\begin{wraptable}[9]{r}{0.42\linewidth}
\vspace{-15pt}
\centering\small
\caption{DINOv2 nearest-neighbor similarity to $\sim$$60$k-image training set.}
\label{tab:nn-memorization}
\setlength{\tabcolsep}{4pt}
\resizebox{\linewidth}{!}{%
\begin{tabular}{lcc}
\toprule
variant & mean NN\,\dn & maximum NN\,\dn \\
\midrule
no-GAN & 0.586 & 0.943 \\
$+$GAN & 0.586 & 0.929 \\
\midrule
$\Delta$ & $+0.0001$ & $-0.014$ \\
\bottomrule
\end{tabular}}
\vspace{-8pt}
\end{wraptable}
\paragraph{Sharpening without mode collapse or memorization.} Unlike a GAN trained as the primary
objective, ours is a post-training term on $\hat{x}_0$ restricted to non-high-noise timesteps that adds high frequency \emph{without}
replacing the backbone, so it sharpens without dropping modes: recall \emph{rises}
($0.36\!\to\!0.41$) rather than falling as adversarial training usually does. To test memorization, we embed each generated image and each of the $\sim$$60$k training images with the frozen
DINOv2 encoder. We then compute cosine similarity to all training images and retain
the largest value as its training-set nearest-neighbor score. Table~\ref{tab:nn-memorization} reports the
mean and maximum of these per-image scores over the evaluation set. The mean changes by only $+0.0001$
(both variants round to $0.586$), while the maximum is lower with the GAN ($0.929$ vs.\ $0.943$). The added
detail is therefore \emph{synthesized}, not copied from nearby training examples. To separate synthesis from generic sharpening, we apply a plain unsharp-mask filter to the no-GAN outputs, tuned to match the $+$GAN HF spectrum (HF log-$\Delta$ and $\alpha$; Appendix Table~\ref{tab:sharpen-control}). Even spectrum-matched sharpening improves FID only to $32.3$ at best, versus $28.59$ for the GAN, and falls short in no-reference quality, ruling out generic sharpening.

\begin{figure}[t]
\begin{center}
\includegraphics[width=0.98\linewidth]{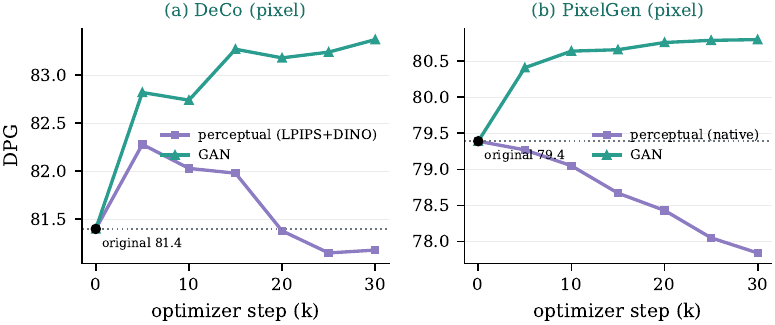}
\end{center}
\caption{DPG Score~\citep{hu2024dpgbench} trajectories initialized from the official, pre-SFT checkpoints. The GAN curves use an image-only DINOv2 discriminator.
(a) DeCo; (b) PixelGen.}
\label{fig:dpg}
\end{figure}

\paragraph{GAN vs.\ perceptual: two routes to visual quality.} The pixel GAN helps because it adds
\emph{real} high frequency, which raises the question of whether adding high frequency by any means is enough.
Perceptual losses (LPIPS $+$ deep DINO features), using the formulation and loss weights adopted by PixelGen, are the standard non-adversarial way to sharpen a converged
model, and, as we show below, they also raise high-frequency power, yet they \emph{hurt}, which makes them the
natural point of comparison. Appendix Table~\ref{tab:perceptual} reports the SFT-checkpoint comparison used in Table~\ref{tab:c1main}: its no-GAN and $+$GAN rows match the main table, with the perceptual arm added. On DeCo, the perceptual arm raises no-reference quality scores (TOPIQ $0.711\!\to\!0.749$, MANIQA $0.636\!\to\!0.661$) but \emph{worsens} FID ($33.3\!\to\!34.1$), pFID ($27.9\!\to\!28.6$), and DPG Score ($81.6\!\to\!81.4$). The main GAN instead improves these metrics to $28.59$, $24.38$, and $83.3$, respectively, while raising TOPIQ and MANIQA to $0.768$ and $0.712$. Thus, its quality gain is real rather than a metric artifact; real photographs themselves score lowest on TOPIQ ($0.57$), showing why no-reference metrics are insufficient on their own. PixelGen follows the same pattern: its native perceptual/SFT baseline has a DPG Score of $78.4$, whereas $+$GAN reaches $80.8$ and wins on every reported metric.
Figure~\ref{fig:dpg} separately reports trajectories initialized directly from the official pre-SFT checkpoints. DeCo changes from $81.4$ to $81.2$ with perceptual supervision and to $83.4$ with GAN, while PixelGen changes from $79.4$ to $77.8$ and $80.8$, respectively.

\begin{table}[t]
\centering
\caption{Image statistics on DPG-Bench~\citep{hu2024dpgbench}. Perceptual supervision reduces color statistics; GAN preserves them while adding sharpness and 2-D HF spectral energy.}
\label{tab:perc-diag}
\resizebox{\columnwidth}{!}{%
\begin{tabular}{lccccc}
\toprule
& Saturation & Contrast & Colorfulness & Sharpness (Lap.\ var) & 2-D HF spectral-energy ratio (\%) \\
\midrule
original & 0.532 & 0.240 & 0.260 & 0.0076 & 4.3 \\
$+$perceptual & 0.511 & 0.210 & 0.221 & 0.022 & 7.9 \\
$+$GAN & 0.544 & 0.239 & 0.256 & 0.052 & 13.9 \\
\bottomrule
\end{tabular}}
\end{table}

\begin{figure}[t]
\centering
\newlength{\pw}\setlength{\pw}{0.150\linewidth}
\definecolor{pgrey}{HTML}{6B6B6B}\definecolor{pcoral}{HTML}{B0642F}\definecolor{pteal}{HTML}{1D7268}
\newcommand{\ph}[2]{\multicolumn{2}{c}{\colorbox{#1}{\parbox[c][1.4ex][c]{1.98\pw}{\centering\scriptsize\textcolor{white}{\textbf{#2}}}}}}
\newcommand{\pii}[1]{\includegraphics[width=\pw]{figs/#1}}
\setlength{\tabcolsep}{1pt}\renewcommand{\arraystretch}{0.35}\setlength{\fboxsep}{0pt}
\begin{tabular}{cccccc}
\ph{pgrey}{Original} & \ph{pcoral}{Perceptual} & \ph{pteal}{$+$GAN}\\
\pii{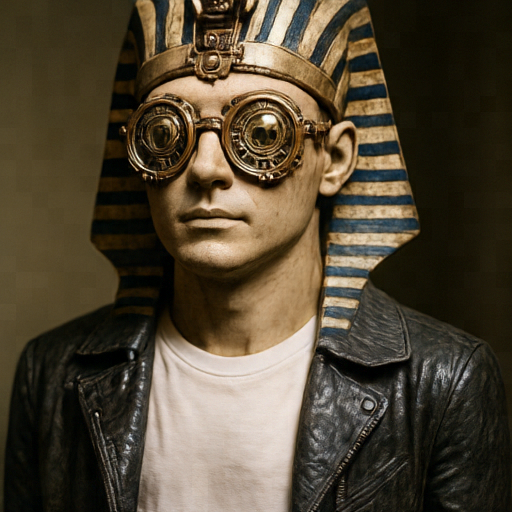}&\pii{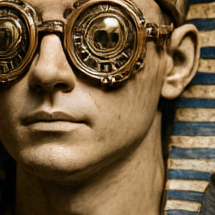}&\pii{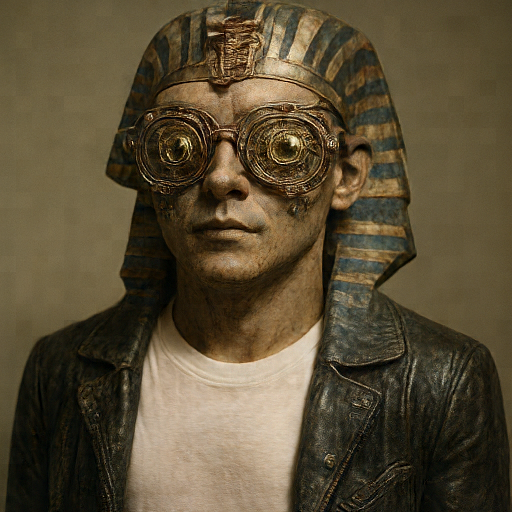}&\pii{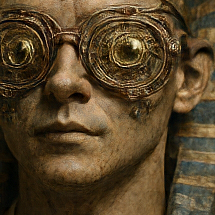}&\pii{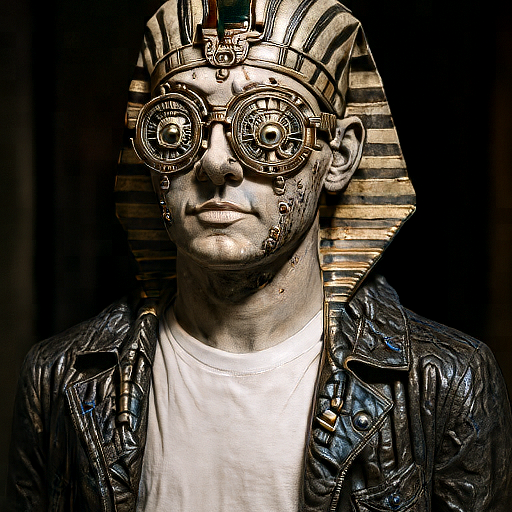}&\pii{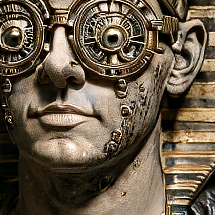}\\[3pt]
\ph{pgrey}{Original} & \ph{pcoral}{Perceptual} & \ph{pteal}{$+$GAN}\\
\pii{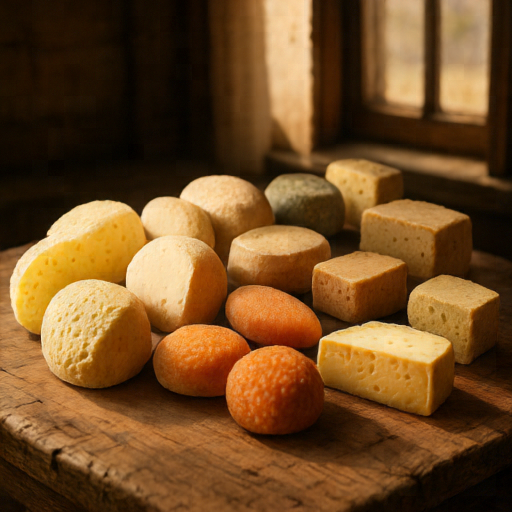}&\pii{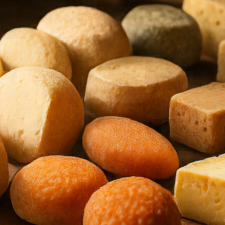}&\pii{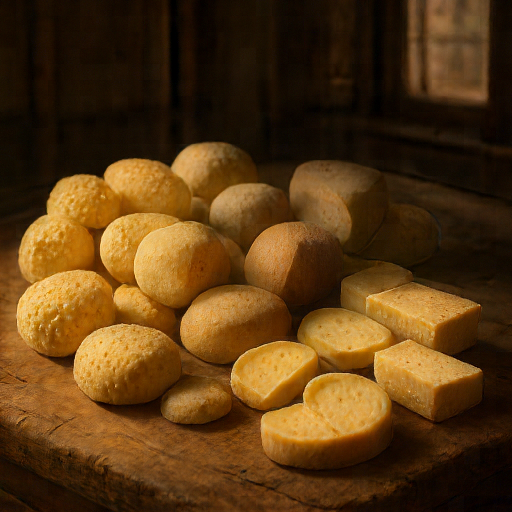}&\pii{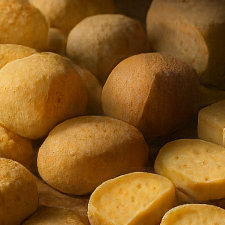}&\pii{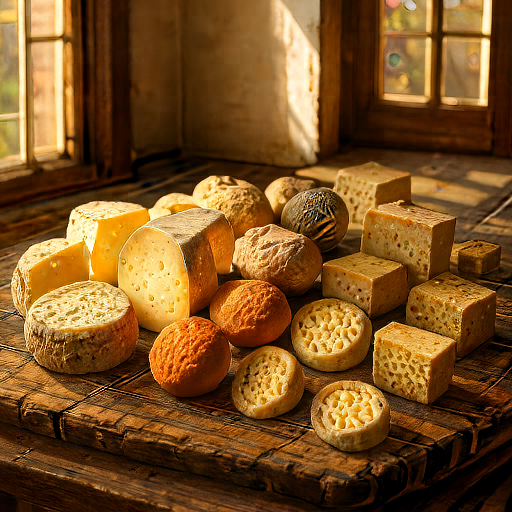}&\pii{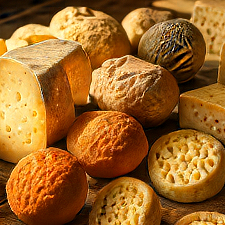}\\[3pt]
\ph{pgrey}{Original} & \ph{pcoral}{Perceptual} & \ph{pteal}{$+$GAN}\\
\pii{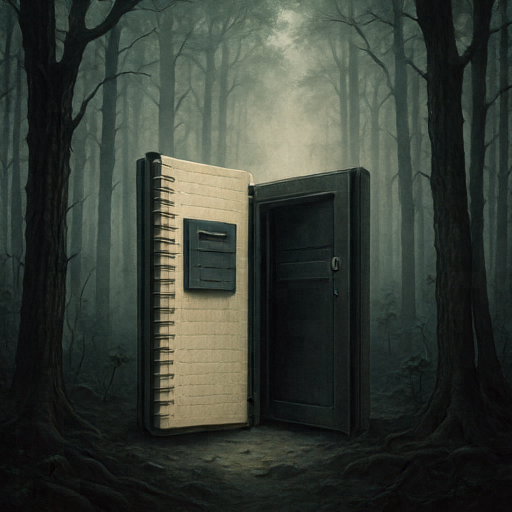}&\pii{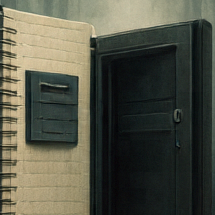}&\pii{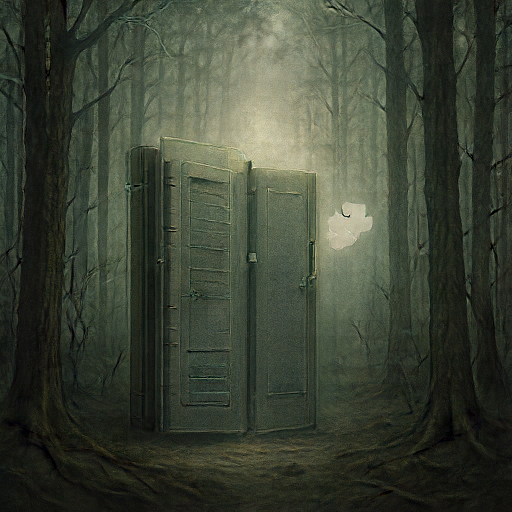}&\pii{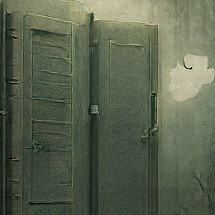}&\pii{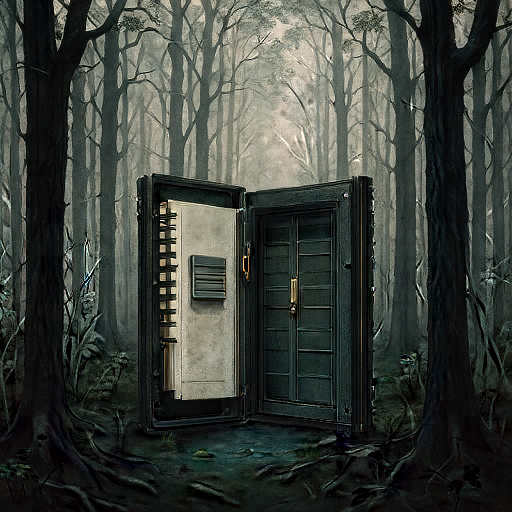}&\pii{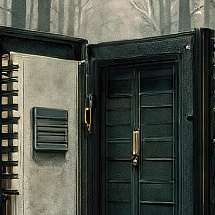}\\
\end{tabular}
\caption{PixelGen outputs (original, perceptual, and $+$GAN; full image $+$ zoom).}
\label{fig:perc-qual}
\end{figure}

\paragraph{Why the perceptual arm hurts: a color-domain shift, not a blur.} Across both the SFT comparison in Appendix Table~\ref{tab:perceptual} and the pre-SFT trajectories in Figure~\ref{fig:dpg}, perceptual supervision weakens DPG Score, whereas GAN improves it. Both objectives add HF and increase Laplacian
sharpness, so the perceptual failure is not over-smoothing. Instead, it consistently lowers saturation,
contrast, and colorfulness, while the GAN preserves them (Table~\ref{tab:perc-diag}). The perceptual loss
therefore buys texture by shifting images toward a desaturated, flattened domain; Figure~\ref{fig:perc-qual}
shows representative cases in which this shift degrades the target while the GAN keeps it sharp and vivid.

% Taken together, the frequency analysis and controls sharpen the empirical claim: adversarial post-training does not merely increase a sharpness score. It restores a systematic natural-image HF deficit while preserving coverage, semantics, and the output distribution. We next test the conditions under which that correction is available.

\section{Pixel--latent contrast and design trade-offs in pixel diffusion}
\label{sec:c2}

Having established the effectiveness and mechanism of adversarial post-training in pixel diffusion, we now
contrast its behavior with latent diffusion. We then examine how discriminator design, timestep gating, and adversarial
weight shape the trade-offs within pixel diffusion.

\subsection{Pixel--latent outcome and spectral contrast}
\label{sec:output-comparison}

\begin{figure}[t]
\centering
\includegraphics[width=0.9\textwidth]{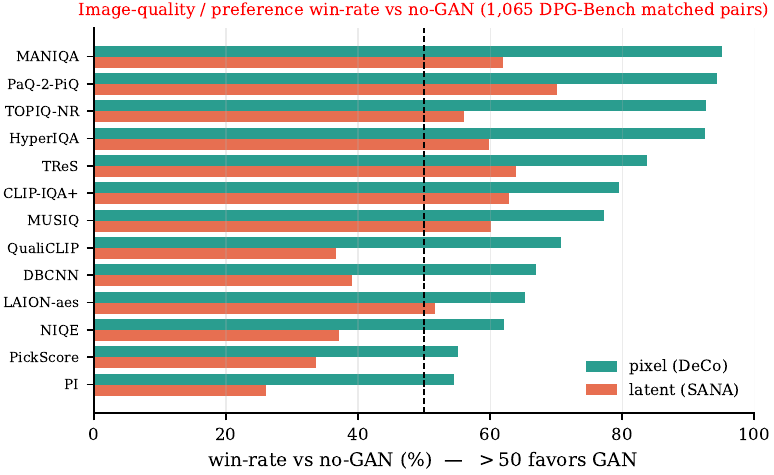}
\caption{\textbf{Image-quality / preference win-rate vs.\ no-GAN} (1,065 DPG-Bench matched pairs~\citep{hu2024dpgbench}). The pixel diffusion's GAN (DeCo) is preferred across metrics, well above the latent diffusion's GAN (SANA).}
\label{fig:winrate}
\end{figure}

We apply adversarial refinement to two latent models: PixArt-$\alpha$ (XL/2, $0.61$B) and SANA ($1.6$B). Relative to the pixel models, the mechanism-level
distinction is the output path: DeCo and PixelGen directly produce RGB, whereas the latent models reach RGB through a frozen decoder. The latent rows in Table~\ref{tab:main} test three discriminator placements. \emph{Self-GAN} uses a copy of
the corresponding generator architecture---SanaMS for SANA and the full DiT for PixArt---as a discriminator
on the predicted clean latent. \emph{Feature-PatchGAN} instead applies a lightweight PatchGAN-style
discriminator directly to SANA's predicted latent. In the \emph{decoded-RGB} variants, the predicted
latent passes through the frozen decoder and the resulting image is scored by either PatchGAN or the same
frozen-DINOv2 discriminator used in the pixel experiments. The VAE remains frozen in every
case; only the diffusion model and discriminator are optimized.

% Temporarily disabled: the quantitative and spectral comparisons below carry this contrast directly.
\iffalse
\begin{figure}[t]
\begin{center}
\includegraphics[width=\linewidth]{figs/fig_overview.pdf}
\end{center}
\caption{\textbf{Adversarial learning directly refines pixel-diffusion outputs.} Each row pairs the architecture (left) with evaluation performance (right; outer $=$ better); DPG Score uses DPG-Bench and the remaining metrics use COCO-30k. \textbf{(b) Pixel diffusion:} the network renders the final RGB image, so the discriminator directly corrects missing detail and improves DeCo across all axes. \textbf{(a) Output-access comparison:} when a frozen VAE mediates the final image, the same supervision yields no comparable joint improvement for PixArt.}
\label{fig:mechanism}
\label{fig:radar}
\end{figure}
\fi

\begin{table}[t]
\caption{\textbf{Pixel--latent comparison.} DPG Score uses DPG-Bench~\citep{hu2024dpgbench}; other metrics use COCO-30k. \best{Bold} marks the better matched result.}
\label{tab:main}
\begin{center}\footnotesize
\setlength{\tabcolsep}{3.5pt}
\resizebox{\textwidth}{!}{%
\begin{tabular}{llcccccccccc}
\toprule
 & variant & DPG Score\,\up & FID\,\dn & IS\,\up & CLIP\,\up & CMMD\,\dn & Rec\,\up & pFID\,\dn & TOPIQ\,\up & MUSIQ\,\up & MANIQA\,\up \\
\midrule
\multirow{2}{*}{\textbf{DeCo} (pixel)}
  & no-GAN SFT & 81.6 & 33.27 & 39.35 & 0.318 & 0.836 & 0.361 & 27.91 & 0.711 & 75.5 & 0.636 \\
  & \;+GAN & \best{83.3} & \best{28.59} & \best{39.77} & \best{0.319} & \best{0.736} & \best{0.406} & \best{24.38} & \best{0.768} & \best{76.5} & \best{0.712} \\
\midrule
\multirow{4}{*}{\textbf{SANA} (latent)}
  & no-GAN SFT (L1) & \best{83.6} & 36.58 & \best{39.20} & 0.320 & 0.872 & \best{0.340} & \best{32.20} & 0.760 & 75.8 & 0.631 \\
  & \;+GAN self-GAN (DiT) & 83.4 & \best{36.24} & 38.75 & \best{0.321} & \best{0.853} & 0.325 & 33.29 & 0.727 & \best{76.2} & 0.614 \\
  & \;+GAN feat-PatchGAN (L2) & 83.5 & 38.41 & 36.41 & 0.320 & 0.915 & 0.334 & 35.18 & \best{0.764} & 76.0 & \best{0.646} \\
  & \;+GAN dec-RGB PatchGAN (B1) & 79.7 & 40.95 & 29.75 & 0.317 & 0.910 & 0.219 & 44.29 & 0.741 & 74.8 & 0.639 \\
\midrule
\multirow{3}{*}{\textbf{PixArt} (latent)}
  & no-GAN SFT & 73.3 & \best{33.70} & \best{38.96} & 0.315 & 0.940 & \best{0.433} & \best{30.07} & \best{0.737} & \best{74.9} & \best{0.599} \\
  & \;+GAN self-GAN (DiT) & \best{75.5} & 35.08 & 38.81 & \best{0.316} & \best{0.892} & 0.425 & 31.46 & 0.660 & 74.3 & 0.534 \\
  & \;+GAN dec-RGB DINOv2 & 72.1 & 34.25 & 32.81 & 0.310 & 0.701 & 0.188 & 36.14 & 0.461 & 64.2 & 0.370 \\
\bottomrule
\end{tabular}}
\end{center}
\end{table}

\begin{wrapfigure}[10]{r}{0.37\linewidth}
\vspace{-20pt}
\centering
\includegraphics[width=\linewidth,trim=320 0 0 0,clip]{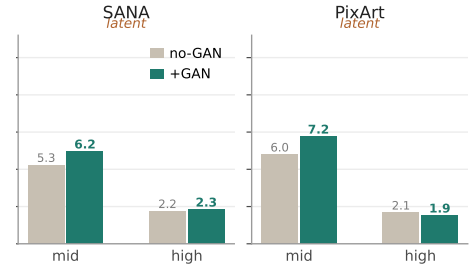}
\vspace{-18pt}
\caption{Frequency response of the self-GAN (DiT) variants in Table~\ref{tab:main}.}
\label{fig:hfbands-latent}
\vspace{-10pt}
\end{wrapfigure}

Table~\ref{tab:main} establishes the outcome contrast under matched no-GAN controls: the tested latent
configurations yield at most isolated metric gains, but none reproduces the joint improvement of DeCo.
Figure~\ref{fig:winrate} transfers the broad image-quality/preference evaluation to matched no-GAN pairs;
the pixel GAN wins across metrics, whereas the latent GAN is substantially weaker and often falls below the
$50\%$ preference threshold. We then transfer the frequency-band and power-law diagnostics from \S\ref{sec:c1}
to decoded latent outputs. Unlike pixel diffusion, neither latent model restores decoded HF: the high-band
share changes only from $2.2\%$ to $2.3\%$ for SANA and from $2.1\%$ to $1.9\%$ for PixArt
(Figure~\ref{fig:hfbands-latent}).

\newpage

\begin{wraptable}[7]{r}{0.37\linewidth}
\vspace{-10pt}
\caption{Spectral statistics of the self-GAN (DiT) variants in Table~\ref{tab:main}.}
\label{tab:latent-spectrum}
\centering\small
\setlength{\tabcolsep}{3.5pt}
\resizebox{\linewidth}{!}{%
\begin{tabular}{lccc}
\toprule
model & HF log-$\Delta$ & $\alpha$ w/o GAN & $\alpha$ $+$GAN \\
\midrule
SANA & $+0.09$ & 2.54 & 2.80 \\
PixArt & $-0.14$ & 2.72 & 2.86 \\
\emph{real} & -- & \multicolumn{2}{c}{$\approx 2.19$} \\
\bottomrule
\end{tabular}}
\vspace{-10pt}
\end{wraptable}
The HF log-power changes and fitted slopes move neither model toward
the natural-image value $\alpha\!\approx\!2.19$ (Table~\ref{tab:latent-spectrum}). Together, the tested latent
models exhibit neither the joint quality gains nor the decoded-HF correction observed in pixel diffusion;
Appendix Figure~\ref{fig:qual} provides same-prompt comparisons, and we test the decoder-mediated
output-access explanation directly below.

\subsection{Frozen VAE decoding attenuates access to high frequencies}

\begin{wrapfigure}{r}{0.44\linewidth}
\vspace{-18pt}
\centering
\includegraphics[width=\linewidth]{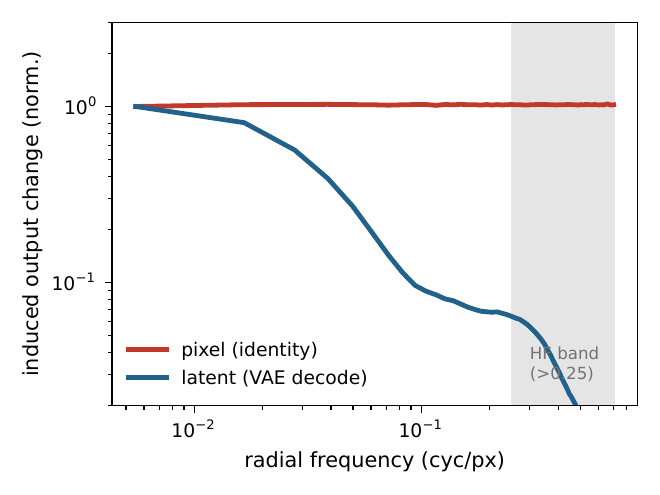}
\vspace{-9pt}
\caption{The frozen PixArt VAE attenuates decoded-HF response by $3.5\text{--}11\times$ relative to the pixel identity map.}
\label{fig:bottleneck}
\vspace{-8pt}
\end{wrapfigure}
What an adversarial loss can change is the generator's native output---pixels in pixel diffusion
and a latent $z$ in latent diffusion. We isolate this representation-to-image constraint with the
output-access operator $A_H=P_HJ_h$: $h$ is the identity in pixel diffusion, whereas PixArt's frozen VAE
decoder mediates the mapping. Under matched native-output perturbations, the VAE produces
$3.5\text{--}11\times$ less decoded-HF response than the pixel identity map (Figure~\ref{fig:bottleneck});
independently, VAE encoding and decoding steepens the real-image spectral slope to $\alpha=2.48$ and removes
approximately $0.2$ dex of HF power (\S\ref{sec:output-comparison}). Gradients are not blocked, but directions
that affect decoded HF are strongly attenuated. Together with PixArt's absent decoded-HF gain and paired
quality results, this provides strong evidence that limited output access is an important mechanism behind the observed
pixel--latent contrast.

\subsection{Trends across discriminator design}
Table~\ref{tab:backbone} reports empirical trends across discriminator designs. In this sweep, pixel
discriminators trained from scratch (PatchGAN, StyleGAN) stay near the no-GAN FID baseline
($32.5$--$33.5$), whereas discriminators built on frozen pretrained backbones (DINOv2, DINOv2-Large,
DINOv3, SigLIP) tend to produce lower FID ($28.6$--$31.0$) and DPG Score up to $83.4$. The individual rows,
however, favor different metrics and do not define a complete ordering. The DINOv2-text main configuration
is reported at $t{\ge}0.53$ and emphasizes FID/CMMD/pFID, whereas the image-only DINOv2 row at
$t{\ge}0.35$ emphasizes a different balance including DPG Score, recall, and NR-IQA.

\subsection{Gate and adversarial weight control the quality trade-off}

\begin{figure}[t]
\centering
\includegraphics[height=5.55cm]{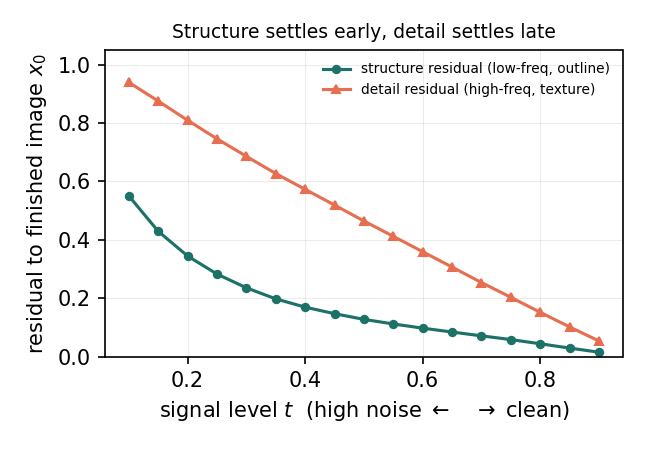}\hspace{0.8em}%
\includegraphics[height=5.55cm]{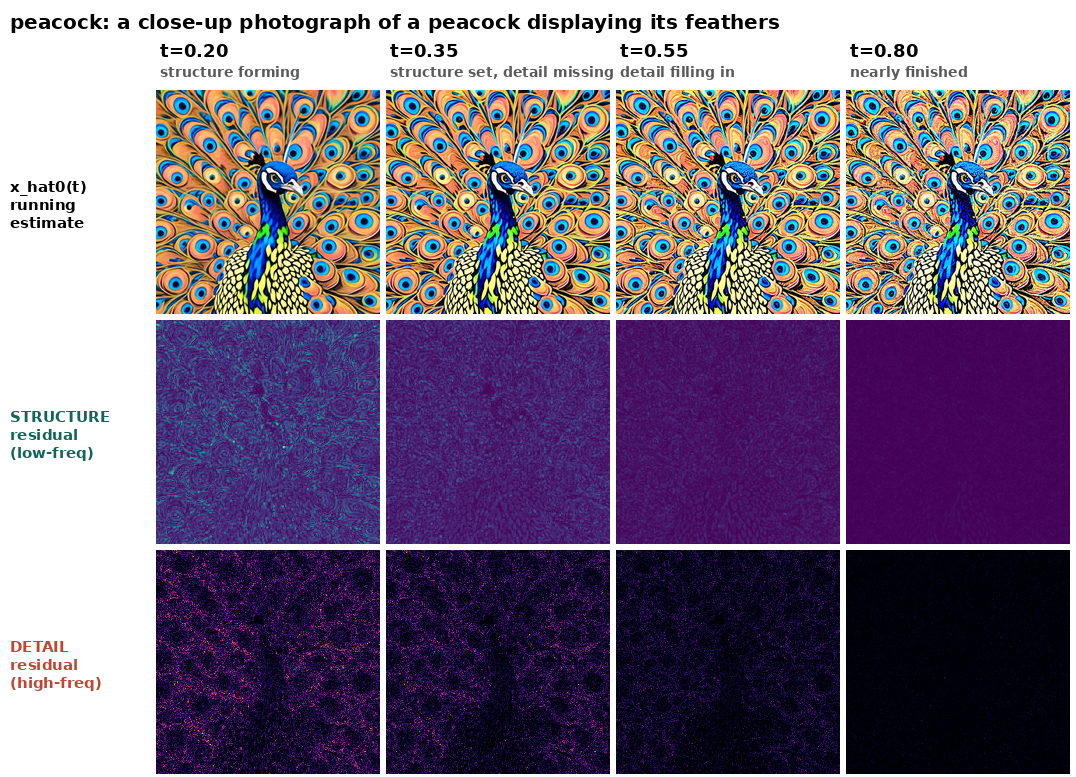}
\caption{Noise-gate rationale. Across $1{,}000$ prompts, structure stabilizes by $t\!\approx\!0.35-0.55$ while detail remains incomplete; per-image residuals show the same coarse-to-fine pattern.}
\label{fig:gate}
\end{figure}

\paragraph{Trends across the noise gate (Table~\ref{tab:deco-range}).} Because the discriminator scores the
predicted clean image $\hat{x}_0$ and the GAN only supplies high-frequency detail, it can help only where two
conditions hold at once: the coarse structure of $\hat{x}_0$ is already settled (otherwise the discriminator
drives detail onto the wrong shapes and corrupts semantics) and the fine detail is still missing (otherwise
there is nothing to add). Since diffusion sampling is coarse-to-fine, these conditions co-occur only in a
specific noise range, which Figure~\ref{fig:gate} makes quantitative: along the sampling trajectory we take
the running clean estimate $\hat{x}_0(t)$ and split its residual to the finished image into a low-frequency
(structure) and a high-frequency (detail) band. Averaged over $1{,}000$ prompts (Fig.~\ref{fig:gate}a), the
structure residual is already low and flattening by $t\!\approx\!0.35$ while the detail residual stays large and
only vanishes near $t\!=\!1$; the per-image residual maps (Fig.~\ref{fig:gate}b) show the same coarse-to-fine
pattern. This gives the gate a clear reading: below it ($t\!\lesssim\!0.3$) the outline is unreliable and the
GAN would push detail onto unformed structure, whereas a very tight gate ($t\!\ge\!0.8$) exposes it to too few
refinement steps. Consequently the sweep yields several useful operating points rather than a single optimum
(Table~\ref{tab:deco-range}): with the fixed DINOv2 discriminator and $w{=}0.1$, $t{\ge}0.35$ emphasizes DPG
Score and no-reference quality, $t{\ge}0.53$ gives higher IS and recall with competitive distribution metrics,
and broader gates favor CMMD and pFID---no threshold dominates the full suite. We therefore use $t{\ge}0.35$
for the fixed-DINOv2 ablations and $t{\ge}0.53$ for the DINOv2-text main configuration.

\paragraph{Trends across GAN weight (Table~\ref{tab:deco-weight}).}
The adversarial weight $w$ is a single strength knob. Raising it monotonically increases no-reference
sharpness, while lower and moderate weights preserve stronger distribution and alignment metrics. No value
dominates across the full metric suite. We use $w{=}0.1$ as the reference operating point for the standard
ablation configuration so that the remaining factors can be compared consistently.

Together, these ablations show how discriminator design, gate, and adversarial weight shift the balance among
fidelity, alignment, coverage, and perceptual quality. They are intended to expose trends rather than identify
a universally optimal configuration or provide a complete conclusion for every metric. We therefore report
them as operating points and leave the configuration choice to the target metric balance.

\section{Conclusion}

We study adversarial post-training as a quality-refinement approach for pretrained pixel diffusion models. Across two backbones, it jointly improves distributional, semantic, and perceptual quality without changing the architecture or inference procedure. Our analysis shows that adversarial supervision restores systematically missing natural-image high-frequency statistics, whereas existing perceptual losses introduce a domain shift. The pixel--latent contrast further suggests that effective refinement depends on direct access to the final image space: frozen decoders attenuate corrections to decoded high-frequency content. These results establish pixel diffusion as a particularly effective setting for adversarial post-training and identify output access as a key factor governing its success.

\bibliographystyle{assets/plainnat}
\bibliography{main}

\clearpage
\appendix
\section{Implementation and model details}
\label{sec:supp}

\paragraph{Implementation and training details.}\label{app:impl}
\emph{Backbones.} DeCo is a $1.1$B-parameter pixel diffusion transformer that denoises directly in
RGB at $512^2$ uses flow matching, and is sampled in $25$ steps. PixelGen
uses a JiT backbone with approximately $1.1$B parameters at $512^2$. The quantitative models are fine-tuned
from their converged public checkpoints on the same public \texttt{blip3o} data.
The DeCo
main-table result instead uses DINOv2-text, which adapts the projection-discriminator conditioning of StyleGAN-T~\citep{sauer2023stylegant}: a pooled Qwen caption embedding is projected into each tapped DINOv2 feature space, and its inner product with the pooled image feature is added to the unconditional patch logits. Unlike StyleGAN-T, whose visual discriminator uses fixed $224{\times}224$ inputs, our DINOv2-text discriminator retains native-resolution scoring and resizes only to the nearest multiple of $14$ ($512{\to}504$). It uses gate $t\ge0.53$; as noted in Table~\ref{tab:backbone}, the
$t\ge0.35$ and $t\ge0.53$ settings are reported as separate operating points with different metric profiles.
\emph{Losses.} Hinge GAN: $\mathcal{L}_G=\lambda\,\mathbb{E}[-D(\hat{x}_0)]$ and
$\mathcal{L}_D=\tfrac12\mathbb{E}[\mathrm{relu}(1-D(x))]+\tfrac12\mathbb{E}[\mathrm{relu}(1+D(\hat{x}_0))]$,
with GAN weight $\lambda=0.1$, applied at non-high-noise timesteps $t\ge0.35$. The original flow-matching
objective is retained, together with a REPA feature-alignment term (weight $0.5$) that aligns denoiser
features to DINOv2 \emph{layer $6$} (distinct from the discriminator's blocks) through a $3$-layer MLP
projection ($1536{\to}1536{\to}768$). DiffAugment (\texttt{color,translation,cutout}) is applied to both real
and fake images before the discriminator and is differentiable, so the gradient reaches the generator.
\emph{Optimization.} Generator: AdamW, lr $10^{-5}$, betas $(0.9,0.999)$, weight decay $0$. Discriminator:
AdamW, lr $2\times10^{-4}$, betas $(0.0,0.99)$, gradient clip $1.0$ (a two-time-scale update; a $10\times$
larger generator lr collapses training). We update the generator first with the discriminator frozen but
retained in the computation graph, then update the discriminator on detached samples using its separate
optimizer. One generator and one discriminator update per step ($1{:}1$), giving a global batch of $512$; EMA decay $0.9999$. \emph{Warm-start, steps, and data.} We fine-tune from the
converged DeCo model for
$30$k steps. The text encoder is Qwen3-1.7B (max length $128$). Training data is
BLIP3o-60k \cite{chen2025blip3o} ($\sim$$60$k image--caption pairs) at $512^2$ with no random crop and training timeshift $4.0$.
At inference we use the AdamLM sampler. PixelGen
$+$GAN follows the same discriminator, loss, and optimizer recipe.

\paragraph{Latent-model pairing and step units.} For each latent backbone, the $+$GAN and no-GAN runs use
the same converged starting checkpoint, model parameterization, data, training horizon, prompts, and
evaluation. PixArt uses a frozen $f8c4$ VAE with eps-prediction, while SANA uses flow matching. PixArt no-GAN
checkpoints are named by micro/dataloader step (optimizer $=$ micro$/4$). All DPG-Bench/COCO-30k comparisons are aligned to a
consistent unit within each figure and table.

\paragraph{Discriminator backbones.}\label{app:disc}
Table~\ref{tab:backbone} compares discriminator designs from two families. All use the same hinge loss and
optimizer described above; the tested gate threshold for each configuration is reported in the table.
\emph{From-scratch discriminators} learn from the RGB image with no pretrained prior.
\textbf{PatchGAN} is the Pix2Pix / \texttt{taming-transformers} $N$-layer discriminator
\citep{isola2017pix2pix}: $4{\times}4$ stride-$2$ convolutions with a patch (rather than whole-image)
receptive field, applied directly to the $[-1,1]$ RGB tensor, with
GroupNorm (batch-statistic-free, so real and fake can be scored in separate DDP passes) and
spectral-normalized convolutions ($\text{ndf}=64$, $3$ layers). \textbf{StyleGAN} is the
ProGAN/StyleGAN discriminator \citep{karras2019stylegan}: equalized-learning-rate convolutions, a fromRGB
$1{\times}1$ conv, a stack of downsampling blocks (two $3{\times}3$ convs $+$ average-pool), a
minibatch-standard-deviation layer at $4{\times}4$, and LeakyReLU$(0.2)$, operating at $256$px.
\emph{Frozen-feature projected discriminators} follow projected GANs \citep{sauer2021projectedgan}: a
\emph{frozen} pretrained backbone supplies features and only lightweight heads are trained.
\textbf{DINOv2} (our main choice) taps patch tokens from blocks $\{2,5,8,11\}$ of a frozen DINOv2 ViT-B/14
\citep{oquab2024dinov2}; each tapped layer has a small head---$1{\times}1$ conv ($768{\to}512$), GroupNorm
($32$ groups), LeakyReLU$(0.2)$, $1{\times}1$ conv ($512{\to}1$)---that emits a per-patch real/fake logit,
and the four layers' logits are concatenated for the hinge loss. Crucially, the image is fed at its
\emph{native} resolution rounded down to a multiple of $14$ ($512\!\to\!504$), \emph{not} bilinearly resized
to a fixed $224$ crop, so the discriminator sees the full $36{\times}36$ patch grid and can score fine detail
across the whole image. \textbf{DINOv2-Large} is the identical design on a frozen ViT-L/14 ($24$ blocks,
width $1024$, taps $\{5,11,17,23\}$); \textbf{DINOv3} swaps in a frozen DINOv3 ViT-B/16
\citep{simeoni2025dinov3}; \textbf{SigLIP} swaps in a frozen SigLIP ViT-B/16 \citep{zhai2023siglip}, whose
$224$-px position embedding is interpolated to the native $32{\times}32$ token grid. \textbf{DINOv2-text}
adds text conditioning to the DINOv2 discriminator in the projection-discriminator style of StyleGAN-T
\citep{sauer2023stylegant}: the pooled Qwen caption embedding is linearly projected (per tapped layer) into
the DINOv2 feature space and added to the patch logits as an inner product with the pooled image feature,
$D(x,y)=D_{\text{uncond}}(x)+\langle V(y),\phi(x)\rangle$, so the score also reflects image--text
consistency. We borrow only this conditioning mechanism: StyleGAN-T uses fixed
$224{\times}224$ visual inputs, whereas our discriminator scores the native image grid, rounded only to a
multiple of $14$ ($512{\to}504$). Every frozen-feature discriminator
keeps the input$\to$feature graph (no stop-gradient), so the generator's adversarial gradient still reaches
the image through the frozen backbone.

\FloatBarrier
\section{Additional quantitative ablations}

\begin{table}[H]
\caption{DeCo discriminator sweep. DPG Score uses DPG-Bench~\citep{hu2024dpgbench}; other metrics use COCO-30k. Gate $t{\ge}0.35$ unless shown; DINOv2-text uses $t{\ge}0.53$. Shading is metric-wise.}
\label{tab:backbone}
\begin{center}\footnotesize
\setlength{\tabcolsep}{3pt}
\resizebox{\textwidth}{!}{%
\begin{tabular}{llcccccccccc}
\toprule
 & discriminator & DPG Score\,\up & FID\,\dn & IS\,\up & CLIP\,\up & CMMD\,\dn & Rec\,\up & pFID\,\dn & TOPIQ\,\up & MUSIQ\,\up & MANIQA\,\up \\
\midrule
 & \emph{no-GAN SFT} & 81.6 & 33.27 & 39.35 & 0.318 & 0.836 & 0.361 & 27.91 & 0.711 & 75.5 & 0.636 \\
\midrule
\multirow{3}{*}{\shortstack[l]{\textbf{no prior}\\(from scratch)}}
  & PatchGAN ($t{\ge}0.53$) & 81.7 & 33.50 & 39.09 & 0.318 & 0.843 & 0.365 & 28.07 & 0.711 & 75.5 & 0.633 \\
  & PatchGAN ($t{\ge}0.20$) & 81.9 & 32.51 & 39.09 & 0.318 & 0.843 & 0.388 & \cellcolor[HTML]{E5EFEC}26.44 & 0.719 & 75.9 & 0.666 \\
  & StyleGAN & 82.1 & 32.93 & 38.84 & 0.318 & 0.834 & 0.365 & 26.73 & 0.695 & 75.5 & 0.603 \\
\midrule
\multirow{5}{*}{\shortstack[l]{\textbf{with prior}\\(frozen feat.)}}
  & DINOv2 & \cellcolor[HTML]{A9CEC6}83.4 & \cellcolor[HTML]{F2F8F6}31.02 & \cellcolor[HTML]{F2F8F6}39.72 & 0.318 & \cellcolor[HTML]{F2F8F6}0.825 & \cellcolor[HTML]{C0DAD3}0.447 & \cellcolor[HTML]{F2F8F6}26.70 & \cellcolor[HTML]{D4E4DF}0.787 & \cellcolor[HTML]{A9CEC6}76.6 & \cellcolor[HTML]{D4E4DF}0.729 \\
  & DINOv2-Large & \cellcolor[HTML]{E5EFEC}83.2 & \cellcolor[HTML]{E5EFEC}30.70 & \cellcolor[HTML]{D4E4DF}40.15 & 0.318 & \cellcolor[HTML]{E5EFEC}0.822 & \cellcolor[HTML]{D4E4DF}0.447 & 26.81 & \cellcolor[HTML]{A9CEC6}0.789 & \cellcolor[HTML]{C0DAD3}76.6 & \cellcolor[HTML]{A9CEC6}0.730 \\
  & DINOv3 & \cellcolor[HTML]{C0DAD3}83.3 & \cellcolor[HTML]{D4E4DF}30.45 & \cellcolor[HTML]{A9CEC6}40.40 & 0.318 & \cellcolor[HTML]{D4E4DF}0.816 & \cellcolor[HTML]{A9CEC6}0.463 & \cellcolor[HTML]{D4E4DF}26.42 & \cellcolor[HTML]{C0DAD3}0.789 & \cellcolor[HTML]{D4E4DF}76.6 & \cellcolor[HTML]{C0DAD3}0.730 \\
  & SigLIP & \cellcolor[HTML]{F2F8F6}83.2 & \cellcolor[HTML]{C0DAD3}30.17 & \cellcolor[HTML]{C0DAD3}40.26 & 0.319 & \cellcolor[HTML]{C0DAD3}0.785 & \cellcolor[HTML]{E5EFEC}0.447 & \cellcolor[HTML]{C0DAD3}25.50 & \cellcolor[HTML]{E5EFEC}0.786 & \cellcolor[HTML]{E5EFEC}76.5 & \cellcolor[HTML]{E5EFEC}0.721 \\
  & DINOv2-text & \cellcolor[HTML]{D4E4DF}83.3 & \cellcolor[HTML]{A9CEC6}28.59 & \cellcolor[HTML]{E5EFEC}39.77 & 0.319 & \cellcolor[HTML]{A9CEC6}0.736 & \cellcolor[HTML]{F2F8F6}0.406 & \cellcolor[HTML]{A9CEC6}24.38 & \cellcolor[HTML]{F2F8F6}0.768 & \cellcolor[HTML]{F2F8F6}76.5 & \cellcolor[HTML]{F2F8F6}0.712 \\
\bottomrule
\end{tabular}}
\end{center}
\end{table}

\begin{table}[H]
\caption{\textbf{DeCo noise-gate sweep} (fixed DINOv2, $w{=}0.1$). DPG Score is evaluated on DPG-Bench~\citep{hu2024dpgbench}; all other metrics use COCO-30k. Different thresholds trade prompt alignment, distribution fidelity, coverage, and no-reference quality; neither $t{\ge}0.35$ nor $t{\ge}0.53$ is uniformly superior.}
\label{tab:deco-range}
\begin{center}\footnotesize
\setlength{\tabcolsep}{3.5pt}
\resizebox{\textwidth}{!}{%
\begin{tabular}{lcccccccccc}
\toprule
gate & DPG Score\,\up & FID\,\dn & IS\,\up & CLIP\,\up & CMMD\,\dn & Rec\,\up & pFID\,\dn & TOPIQ\,\up & MUSIQ\,\up & MANIQA\,\up \\
\midrule
$t{\ge}0.0$ (full) & 82.6 & 31.63 & 37.91 & 0.319 & 0.755 & 0.427 & 26.12 & 0.767 & 76.56 & 0.727 \\
$t{\ge}0.20$ & 83.1 & 31.16 & 38.93 & 0.318 & 0.798 & 0.427 & 26.16 & 0.778 & 76.63 & 0.724 \\
$t{\ge}0.35$ & 83.4 & 31.02 & 39.72 & 0.318 & 0.825 & 0.447 & 26.70 & 0.787 & 76.58 & 0.729 \\
$t{\ge}0.53$ & 82.9 & 31.20 & 41.40 & 0.319 & 0.799 & 0.450 & 26.55 & 0.781 & 76.36 & 0.715 \\
$t{\ge}0.80$ & 82.3 & 31.99 & 40.38 & 0.319 & 0.811 & 0.412 & 26.98 & 0.766 & 76.01 & 0.694 \\
\bottomrule
\end{tabular}}
\end{center}
\end{table}

\begin{table}[H]
\caption{\textbf{DeCo GAN-weight sweep} (DINOv2, gate $t{\ge}0.35$). DPG Score is evaluated on DPG-Bench~\citep{hu2024dpgbench}; all other metrics use COCO-30k.}
\label{tab:deco-weight}
\begin{center}\footnotesize
\setlength{\tabcolsep}{3.5pt}
\resizebox{\textwidth}{!}{%
\begin{tabular}{lcccccccccc}
\toprule
weight & DPG Score\,\up & FID\,\dn & IS\,\up & CLIP\,\up & CMMD\,\dn & Rec\,\up & pFID\,\dn & TOPIQ\,\up & MUSIQ\,\up & MANIQA\,\up \\
\midrule
$w=0.01$ & 82.4 & 31.71 & 40.59 & 0.318 & 0.798 & 0.420 & 26.32 & 0.769 & 76.44 & 0.715 \\
$w=0.05$ & 82.9 & 30.93 & 40.31 & 0.318 & 0.805 & 0.447 & 26.05 & 0.777 & 76.51 & 0.721 \\
$w=0.10$ & 83.4 & 31.02 & 39.72 & 0.318 & 0.825 & 0.447 & 26.70 & 0.787 & 76.58 & 0.729 \\
$w=0.50$ & 83.0 & 30.73 & 38.65 & 0.318 & 0.846 & 0.436 & 27.80 & 0.798 & 76.94 & 0.736 \\
$w=1.00$ & 82.9 & 30.73 & 38.08 & 0.318 & 0.866 & 0.424 & 28.97 & 0.804 & 76.98 & 0.749 \\
\bottomrule
\end{tabular}}
\end{center}
\end{table}

\paragraph{Full GAN vs.\ perceptual comparison.} Table~\ref{tab:perceptual} gives the complete
GAN-vs-perceptual comparison on both pixel backbones, including the perceptual (LPIPS$+$DINO) arm omitted
from the main-text Table~\ref{tab:c1main}. On DeCo, the perceptual arm raises no-reference sharpness (TOPIQ,
MANIQA) but worsens the reference-based distribution and alignment metrics (FID, pFID, DPG Score), whereas the
main DINOv2-text GAN improves both at once. PixelGen tells the same story. For PixelGen, ``perceptual (native)'' is the native perceptual/SFT baseline, and the $+$GAN variant replaces that perceptual supervision with the adversarial loss rather than adding GAN on top of it.

\begin{table}[H]
\caption{\textbf{GAN vs.\ perceptual fine-tuning on two pixel backbones.} DPG Score is evaluated on DPG-Bench~\citep{hu2024dpgbench}; all other metrics use COCO-30k. The perceptual arm raises no-reference sharpness but trades away distribution metrics; the GAN improves both. The DeCo GAN row uses the main DINOv2-text configuration. \best{Bold} marks the favorable value per column within each model.}
\label{tab:perceptual}
\begin{center}\footnotesize
\setlength{\tabcolsep}{3.5pt}
\resizebox{\textwidth}{!}{%
\begin{tabular}{llccccccccc}
\toprule
 & variant & DPG Score\,\up & FID\,\dn & IS\,\up & CMMD\,\dn & Rec\,\up & pFID\,\dn & TOPIQ\,\up & MUSIQ\,\up & MANIQA\,\up \\
\midrule
\multirow{3}{*}{\textbf{DeCo}}
  & no-GAN SFT & 81.6 & 33.27 & 39.35 & 0.836 & 0.361 & 27.91 & 0.711 & 75.5 & 0.636 \\
  & perceptual (LPIPS+DINO) & 81.4 & 34.14 & 38.95 & 0.826 & 0.367 & 28.64 & 0.749 & 76.2 & 0.661 \\
  & \;$+$GAN (DINOv2-text) & \best{83.3} & \best{28.59} & \best{39.77} & \best{0.736} & \best{0.406} & \best{24.38} & \best{0.768} & \best{76.5} & \best{0.712} \\
\midrule
\multirow{2}{*}{\textbf{PixelGen}}
  & perceptual (native) & 78.4 & 33.94 & 38.58 & 0.762 & 0.319 & 30.81 & 0.755 & 75.7 & 0.645 \\
  & \;+GAN & \best{80.8} & \best{33.20} & \best{38.76} & \best{0.725} & \best{0.403} & \best{30.51} & \best{0.799} & \best{77.0} & \best{0.723} \\
\bottomrule
\end{tabular}}
\end{center}
\end{table}

\FloatBarrier
\section{Additional qualitative comparisons}

\begin{figure}[H]
\centering
\newlength{\qw}\setlength{\qw}{0.096\linewidth}
\definecolor{qgrey}{HTML}{6B6B6B}\definecolor{qcoral}{HTML}{B0642F}\definecolor{qteal}{HTML}{1D7268}
\newcommand{\qh}[2]{\colorbox{#1}{\parbox[c]{\qw}{\centering\tiny\textcolor{white}{#2}\strut}}}
\newcommand{\qi}[1]{\includegraphics[width=\qw]{figs/#1}}
\setlength{\tabcolsep}{1pt}\renewcommand{\arraystretch}{0.35}\setlength{\fboxsep}{0pt}
\begin{tabular}{cccc@{\hspace{5pt}}cccc}
\qh{qgrey}{PixArt\\no-GAN} & \qh{qcoral}{PixArt\\+GAN} & \qh{qgrey}{DeCo\\no-GAN} & \qh{qteal}{DeCo\\+GAN} &
\qh{qgrey}{PixArt\\no-GAN} & \qh{qcoral}{PixArt\\+GAN} & \qh{qgrey}{DeCo\\no-GAN} & \qh{qteal}{DeCo\\+GAN}\\
\multicolumn{4}{c}{\scriptsize full image} & \multicolumn{4}{c}{\scriptsize center zoom}\\
\qi{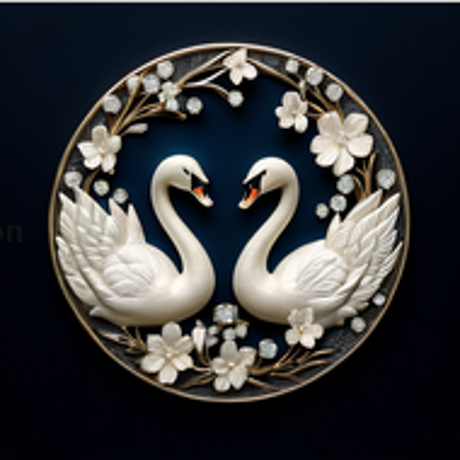}&\qi{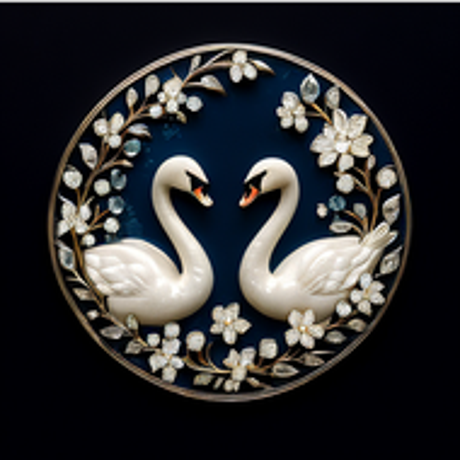}&\qi{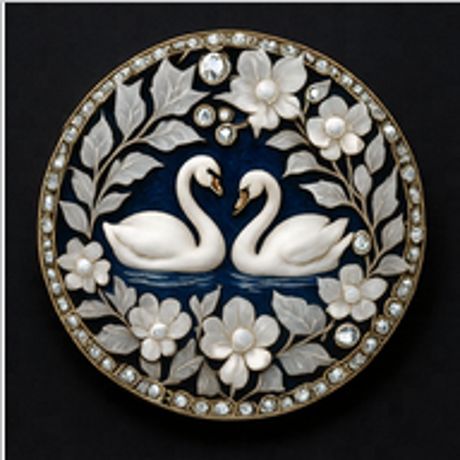}&\qi{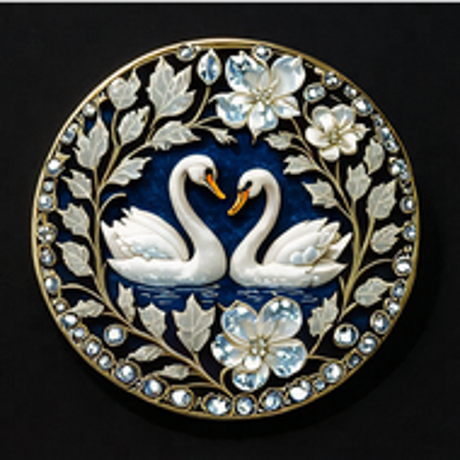}&\qi{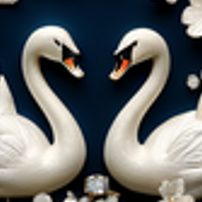}&\qi{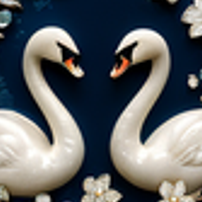}&\qi{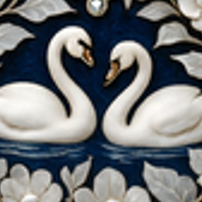}&\qi{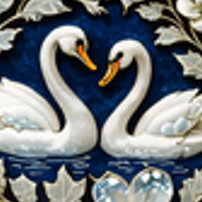}\\[2pt]
\qi{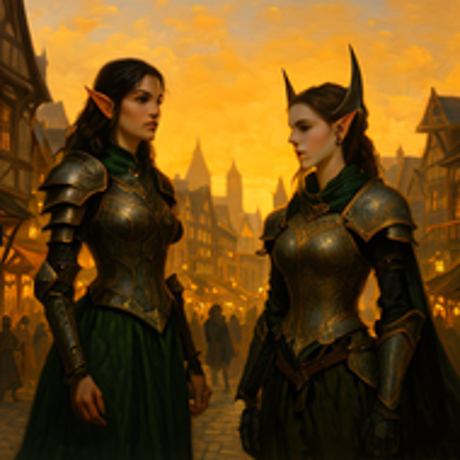}&\qi{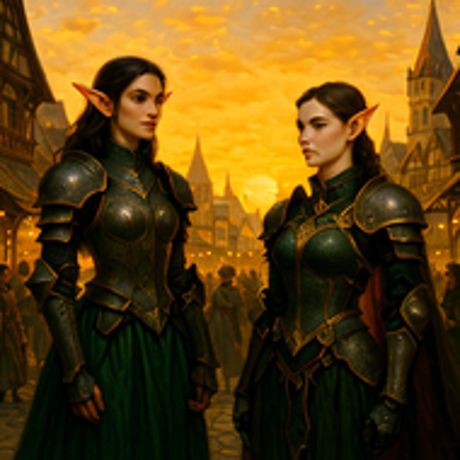}&\qi{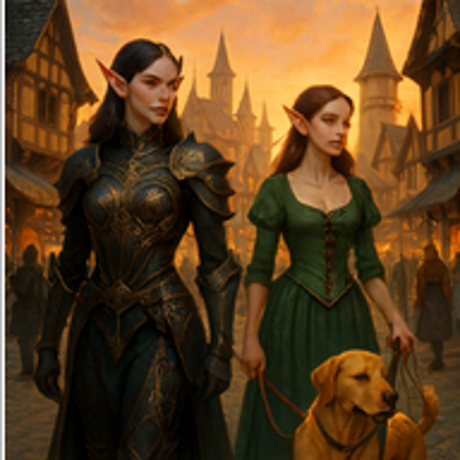}&\qi{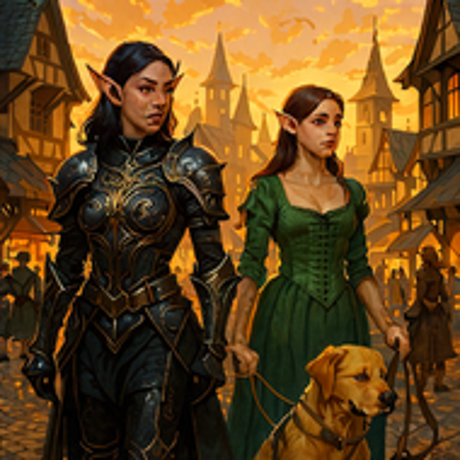}&\qi{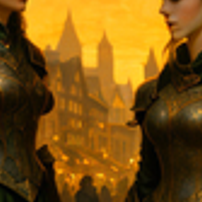}&\qi{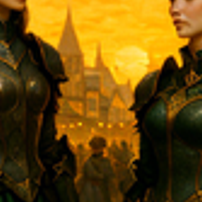}&\qi{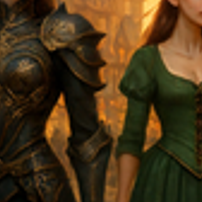}&\qi{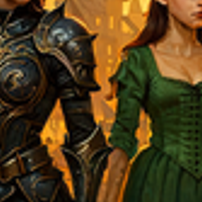}\\[2pt]
\qi{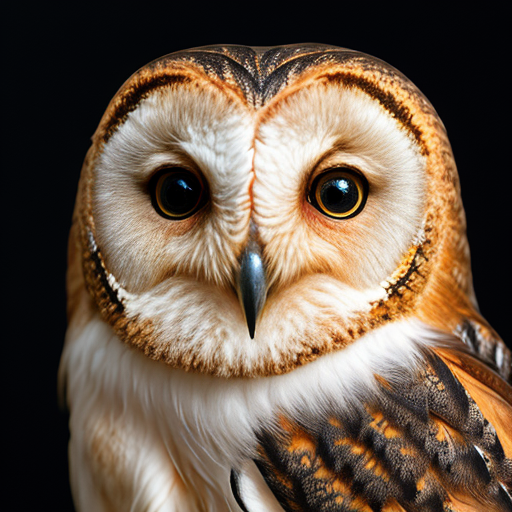}&\qi{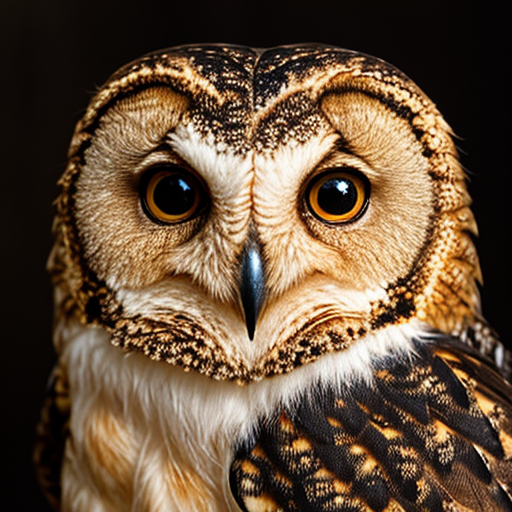}&\qi{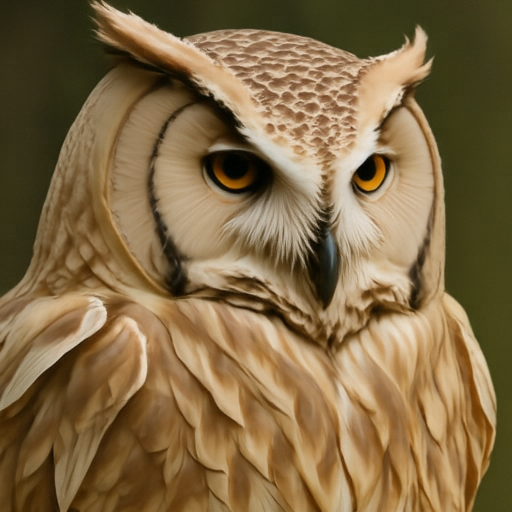}&\qi{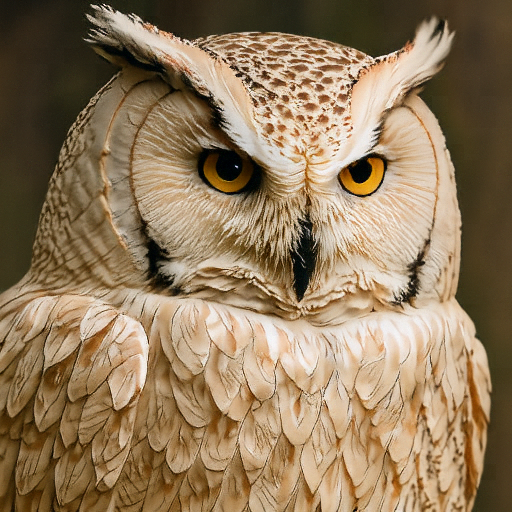}&\qi{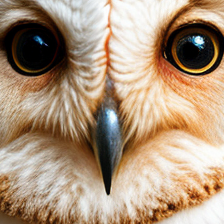}&\qi{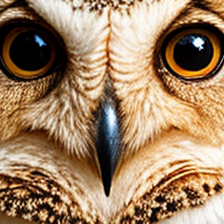}&\qi{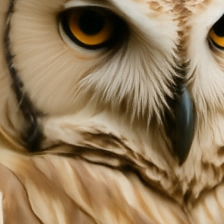}&\qi{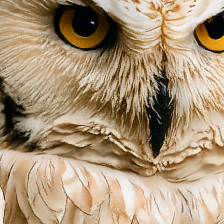}\\
\end{tabular}
\caption{\textbf{Same-prompt output-access comparison} (three prompts; full image $+$ center zoom). PixArt no-GAN vs.\ $+$GAN are nearly identical, whereas DeCo $+$GAN visibly gains fine detail.}
\label{fig:qual}
\end{figure}

\begin{figure}[p]
\centering
\includegraphics[width=\textwidth]{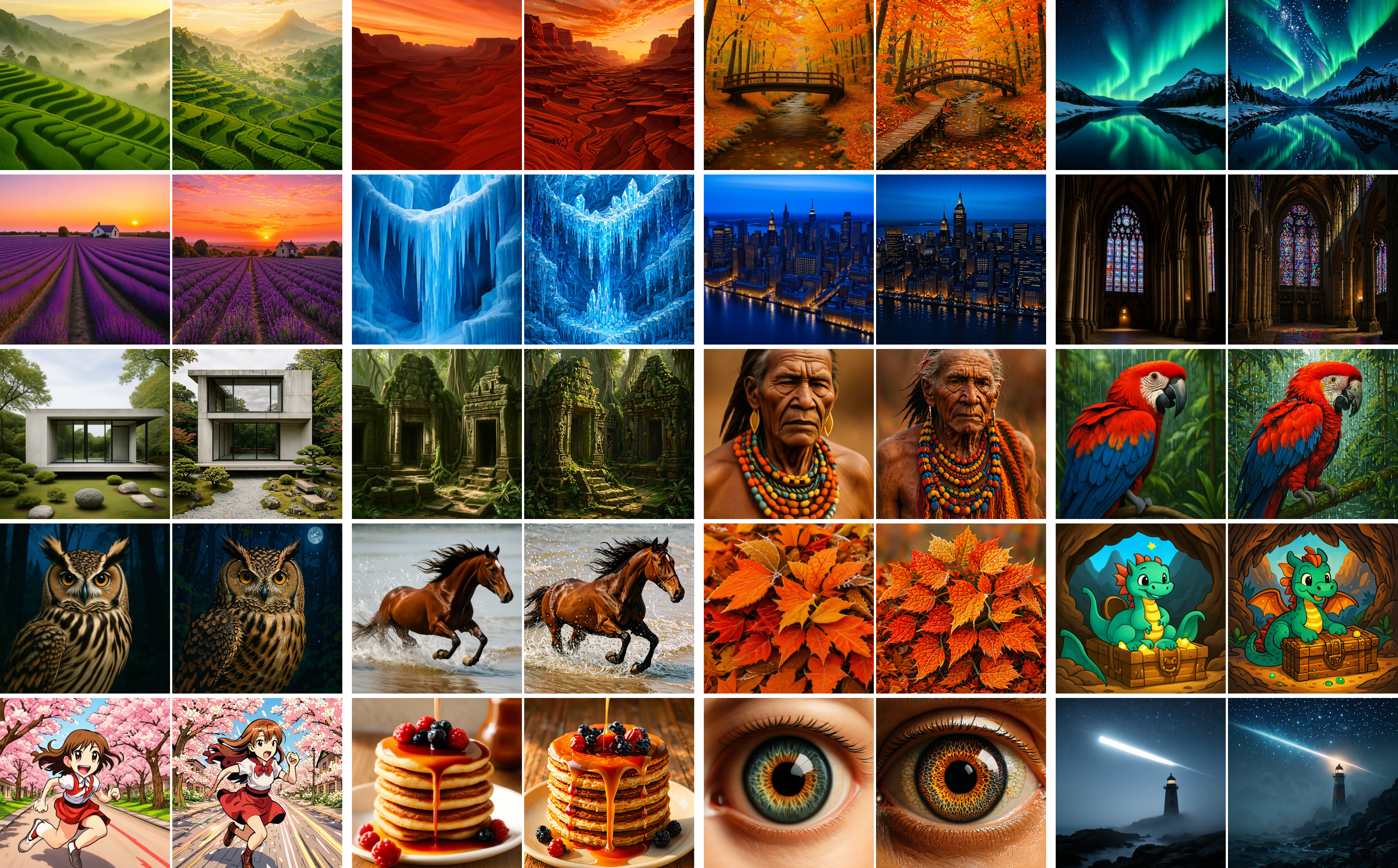}
\caption{\textbf{Additional $512$px comparisons.} Each adjacent pair: \emph{left} without GAN,
\emph{right} with our GAN fine-tuning (DeCo, same prompt and seed). Best viewed zoomed in.}
\label{fig:supp-a}
\end{figure}

Figure~\ref{fig:qual} compares pixel and latent outputs
under the same prompts, with full images and center crops. Figures~\ref{fig:supp-a} shows further no-GAN vs.\ $+$GAN pairs
from DeCo at $512^2$ across a wide range of prompts and styles (landscapes, cities, architecture, people,
animals, food, macro, night, and stylized art). In each adjacent pair the left image is without the GAN and
the right image is with our GAN fine-tuning, generated from the same prompt and seed.

\FloatBarrier
\section{Guidance and sampler controls}

\paragraph{Spectral-measure conventions.}
The three percentage-valued spectral diagnostics use different normalizations and evaluation
sets and therefore should not be compared numerically. Figure~\ref{fig:hfbands} reports the radial-profile
band share on COCO-30k: after DC removal and Hann windowing, the 2-D power spectrum is azimuthally averaged,
and the power in each radial interval is normalized by the total radial-profile power. Table~\ref{tab:perc-diag}
reports the 2-D HF spectral-energy ratio on the $1{,}065$ DPG-Bench prompts: Fourier energy at
$f>0.25$ cyc/px is divided by total 2-D spectral energy, so outer radii receive more weight because they
contain more Fourier coefficients. The CFG/order control below reports decoded radial-power shares on its
own $300$ fixed prompts and is intended only for comparisons among the settings within that control, not for
direct numerical comparison with Figure~\ref{fig:hfbands} or Table~\ref{tab:perc-diag}.

\begin{figure}[h]
\centering
\includegraphics[width=\textwidth]{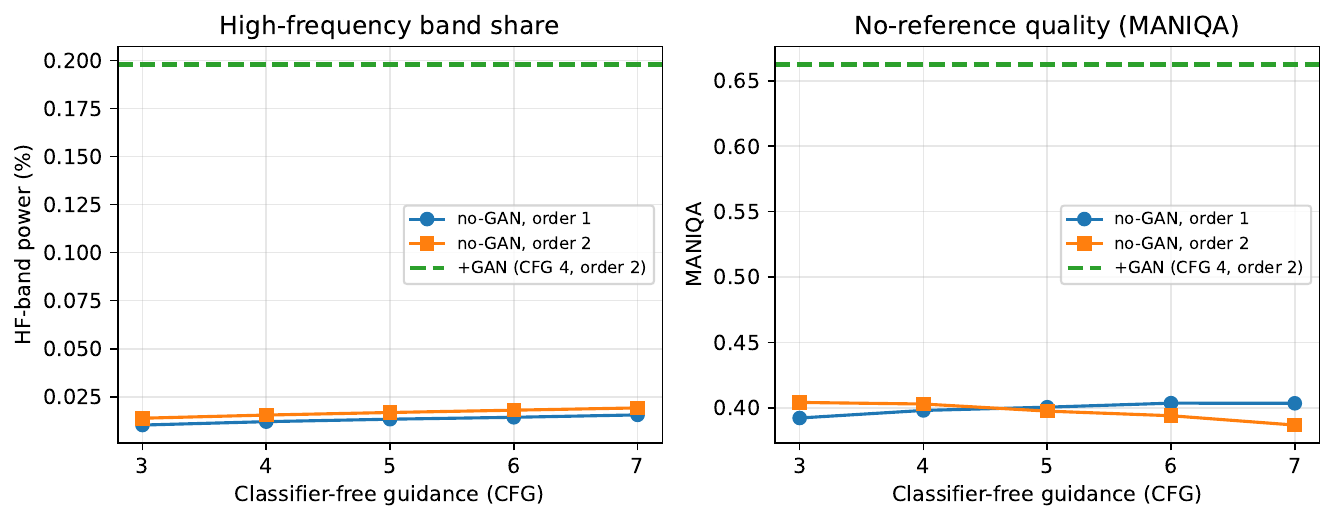}
\caption{\textbf{Guidance and sampler order do not recover the GAN's high frequency.} No-GAN model swept over
CFG and sampler order (orders $1$ and $2$); the $+$GAN model (CFG $4$, order $2$) is the green dashed
reference. Left: decoded high-frequency band share; right: MANIQA. Both no-GAN curves stay far below $+$GAN.}
\label{fig:cfg-order}
\end{figure}

A natural question is whether the GAN's sharpness can instead be obtained from the \emph{no-GAN} model by
raising classifier-free guidance (CFG) or by using a higher-order sampler. It cannot. On a fixed set of
$300$ prompts (identical prompts and seeds across settings, $512^2$, $25$-step AdamLM), we sweep the no-GAN
model over CFG $\in\{3,4,5,6,7\}$ and sampler order $\in\{1,2\}$ and compare against the $+$GAN model at its
default (CFG $4$, order $2$); Table~\ref{tab:cfg-order} and Figure~\ref{fig:cfg-order} report the decoded
high-frequency band share (our spectrum pipeline), no-reference quality (TOPIQ, MANIQA), and
mean HSV saturation. The $+$GAN model carries $0.198\%$ of its power in the high band, whereas the best
no-GAN setting reaches only $0.019\%$ (CFG $7$, order $2$)---a $\sim$$10\times$ gap that no guidance or order
setting closes, with the mid band showing the same pattern. Raising CFG does not help and actively hurts: it
monotonically increases saturation ($0.527\!\to\!0.590$), i.e.\
oversaturation, while no-reference quality falls at the high end (order-$2$ MANIQA drops from $0.404$ to
$0.387$). Switching from order $1$ to order $2$ barely moves the high band. On no-reference quality the GAN
is in a different regime entirely (MANIQA $0.66$ vs.\ $\le0.40$; TOPIQ $0.73$ vs.\ $\le0.52$). The added
high-frequency detail is thus a property of the adversarial fine-tuning, not something recoverable by
inference-time tuning.

\begin{table}[h]
\caption{\textbf{No-GAN CFG/order sweep vs.\ the $+$GAN model} (DeCo, $300$ fixed prompts/seeds). HF/mid bands are the decoded radial-power shares, and TOPIQ/MANIQA are no-reference quality metrics. Neither higher CFG nor order-$2$ sampling approaches the $+$GAN high-frequency or quality.}
\label{tab:cfg-order}
\begin{center}\footnotesize
\setlength{\tabcolsep}{5pt}
\begin{tabular}{llccccc}
\toprule
model & CFG & order & HF\,\% & mid\,\% & TOPIQ\,\up & MANIQA\,\up \\
\midrule
no-GAN & 3 & 1 & 0.010 & 0.190 & 0.502 & 0.392 \\
no-GAN & 3 & 2 & 0.014 & 0.221 & 0.520 & 0.404 \\
no-GAN & 4 & 1 & 0.012 & 0.204 & 0.510 & 0.398 \\
no-GAN & 4 & 2 & 0.015 & 0.234 & 0.516 & 0.403 \\
no-GAN & 5 & 1 & 0.013 & 0.215 & 0.512 & 0.401 \\
no-GAN & 5 & 2 & 0.017 & 0.245 & 0.505 & 0.398 \\
no-GAN & 6 & 1 & 0.014 & 0.223 & 0.516 & 0.404 \\
no-GAN & 6 & 2 & 0.018 & 0.256 & 0.498 & 0.394 \\
no-GAN & 7 & 1 & 0.016 & 0.231 & 0.515 & 0.404 \\
no-GAN & 7 & 2 & 0.019 & 0.255 & 0.485 & 0.387 \\
\midrule
\textbf{$+$GAN} & \textbf{4} & \textbf{2} & \textbf{0.198} & \textbf{0.594} & \textbf{0.730} & \textbf{0.662} \\
$+$GAN & 4 & 1 & 0.160 & 0.515 & 0.746 & 0.664 \\
\bottomrule
\end{tabular}
\end{center}
\end{table}

\paragraph{Spectrum-matched sharpening control.} A natural concern is that any operation that raises high-frequency power would reproduce our results. It does not. We apply a plain unsharp-mask filter to the no-GAN DeCo outputs and tune its strength to match the $+$GAN change in either high-frequency band power (HF log-$\Delta$) or spectral slope ($\alpha$). As Table~\ref{tab:sharpen-control} shows, the filter reaches the GAN's spectral signature (and even a small FID improvement), but recovers only a small fraction of the GAN's FID gain and does not approach its no-reference quality---so the adversarial improvement is not generic sharpening.

\begin{table}[h]
\centering\small
\caption{\textbf{Spectrum-matched sharpening control} (COCO-30k). A plain unsharp-mask filter applied to the no-GAN outputs, tuned to match the $+$GAN high-frequency spectrum (HF log-$\Delta$ and slope $\alpha$), reproduces the GAN's spectral signature but recovers only a small fraction of its FID gain and does not reach its no-reference quality. The no-GAN and $+$GAN reference rows match Tables~\ref{tab:c1main} and~\ref{tab:spectrum-app}.}
\label{tab:sharpen-control}
\begin{tabular}{lccccc}
\toprule
method & HF log-$\Delta$ & $\alpha$ & FID $\downarrow$ & TOPIQ $\uparrow$ & MANIQA $\uparrow$ \\
\midrule
no-GAN & $0.00$ & $2.59$ & $33.27$ & $0.711$ & $0.636$ \\
$+$unsharp (HF-matched) & $+0.35$ & $2.38$ & $32.7$ & $0.719$ & $0.673$ \\
$+$unsharp (slope-matched) & $+0.55$ & $2.23$ & $32.3$ & $0.691$ & $0.669$ \\
\midrule
$+$GAN (main, Table~\ref{tab:c1main}) & $+0.34$ & $2.24$ & $\textbf{28.59}$ & $\textbf{0.768}$ & $\textbf{0.712}$ \\
\bottomrule
\end{tabular}
\end{table}

\end{document}